\documentclass[10pt,journal,compsoc]{IEEEtran}
\ifCLASSOPTIONcompsoc
\usepackage[nocompress]{cite}
\else
\usepackage{cite}
\fi

\usepackage{graphicx}
\usepackage{amsmath}
\usepackage{amssymb}
\usepackage{color}
\usepackage{epstopdf}
\usepackage{array}
\usepackage{multirow}
\usepackage{amsfonts}
\usepackage{booktabs}
\usepackage{float}
\usepackage{subfigure}
\usepackage{algorithm}
\usepackage{algpseudocode}
\usepackage[switch]{lineno}

\begin{document}
%
\title{Knowledge-Guided Multi-Label Few-Shot Learning for General Image Recognition}

\author{Tianshui Chen, Liang Lin, Riquan Chen, Xiaolu Hui, and Hefeng Wu
\IEEEcompsocitemizethanks{\IEEEcompsocthanksitem This work was supported in part by the National Key Research and Development Program of China under Grant No. 2018YFC0830103, in part by National Natural Science Foundation of China (NSFC) under Grant No. 61876045 and 61836012, and in part by Zhujiang Science and Technology New Star Project of Guangzhou under Grant No. 201906010057. (Corresponding author: Liang Lin).

T. Chen and L. Lin are with the School of Data and Computer Science, Sun Yat-sen University and DarkMatter AI Research, Guangzhou, China. R. Chen, X. Hui, and H. Wu are with the School of Data and Computer Science, Sun Yat-sen University, Guangzhou, China (Email: tianshuichen@gmail.com, linliang@ieee.org, \{chenrq6, huixlu\}@mail2.sysu.edu.cn, wuhefeng@gmail.com). \protect\\
}
}


\markboth{IEEE Transactions on Pattern Analysis and Machine Intelligence}%
{Chen \MakeLowercase{\textit{et al.}}: Knowledge-Guided Multi-Label Few-Shot Learning for General Image Recognition}

\IEEEtitleabstractindextext{
\begin{abstract}
Recognizing multiple labels of an image is a practical yet challenging task, and remarkable progress has been achieved by searching for semantic regions and exploiting label dependencies. However, current works utilize RNN/LSTM to implicitly capture sequential region/label dependencies, which cannot fully explore mutual interactions among the semantic regions/labels and do not explicitly integrate label co-occurrences. In addition, these works require large amounts of training samples for each category, and they are unable to generalize to novel categories with limited samples. To address these issues, we propose a knowledge-guided graph routing (KGGR) framework, which unifies prior knowledge of statistical label correlations with deep neural networks. The framework exploits prior knowledge to guide adaptive information propagation among different categories to facilitate multi-label analysis and reduce the dependency of training samples. Specifically, it first builds a structured knowledge graph to correlate different labels based on statistical label co-occurrence. Then, it introduces the label semantics to guide learning semantic-specific features to initialize the graph, and it exploits a graph propagation network to explore graph node interactions, enabling learning contextualized image feature representations. Moreover, we initialize each graph node with the classifier weights for the corresponding label and apply another propagation network to transfer node messages through the graph. In this way, it can facilitate exploiting the information of correlated labels to help train better classifiers, especially for labels with limited training samples. We conduct extensive experiments on the traditional multi-label image recognition (MLR) and multi-label few-shot learning (ML-FSL) tasks and show that our KGGR framework outperforms the current state-of-the-art methods by sizable margins on the public benchmarks.
\end{abstract}

}


\maketitle

\IEEEdisplaynontitleabstractindextext

%
\IEEEpeerreviewmaketitle

\IEEEraisesectionheading{\section{Introduction}
\label{sec:introduction}}
\IEEEPARstart{R}{eal-world} images generally contain objects belonging to multiple diverse categories; thus, recognizing the multiple object categories in images is a more fundamental and practical task compared with single-object image recognition in computer vision and multimedia fields. Recently, researchers have concentrated on developing a series of algorithms for multi-label image analysis \cite{li2017improving,wei2016hcp,cabral2014matrix,song2014generalized,lapin2017analysis} that underpin many critical applications, such as content-based image retrieval and recommendation systems \cite{chua1994concept,yang2015pinterest}. Despite these achievements, identifying the existence of multiple semantic categories requires not only mining diverse semantic object regions but also capturing the interplay among these regions and their semantics, making multi-label image analysis a challenging and unsolved task.

The current MLR approaches mainly employ object localization techniques \cite{wei2016hcp,yang2016exploit,ren2017faster} or adopt visual attentional mechanisms \cite{zhu2017learning,wang2017multi} to adaptively discover regions with meaningful semantic categories. However, object localization techniques \cite{uijlings2013selective,zitnick2014edge,zhang2017sequential,pont2016multiscale} need to search numerous category-agnostic and superfluous region proposals, which cannot be unified into deep neural networks for end-to-end optimization. Moreover, visual attentional mechanisms cannot accurately locate semantic object regions due to a lack of explicit supervision and guidance. More recent works have further introduced RNN/LSTM models \cite{hochreiter1997long,wang2016cnn,chen2018recurrent} to identify the contextual dependencies among located regions and thus implicitly capture label dependencies. However, these algorithms merely model sequential region/label dependencies; they cannot fully exploit these properties because a direct correlation or dependency exists between each region/label pair. In addition, they do not explicitly integrate statistical category correlations that provide direct and key guidance to aid multi-label image analysis.

The current methods depend on deep convolution networks \cite{he2016deep,simonyan2015very} for learning image features; such networks require large numbers of training samples for each category and are unable to generalize to novel categories with limited samples. To address this issue, researchers have recently developed a series of few-shot learning algorithms that can learn novel categories after being trained on a set of base categories with sufficient training samples. These algorithms utilize the meta-learning paradigm \cite{sung2018learning,kim2019variational} to help learn the novel categories by distilling the knowledge of the categories with sufficient training samples or use sample synthesis technologies \cite{hariharan2017low} to generate more diverse samples for the novel categories. Although these techniques have achieved impressive progress, they focus mainly on single-label scenarios rather than on more general multi-label cases.

Objects in visual scenes commonly have strong correlations. For example, desks tend to co-occur with chairs, while computers usually co-exist with keyboards. These correlations can provide extra guidance for capturing the interplay among different categories and thus facilitate multi-label analysis, especially for complex and few-shot scenarios. In this work, we show that semantic correlations can be explicitly represented by a structured knowledge graph and that such interplay can be effectively captured by information propagation through the graph. To this end, we propose a knowledge-guided graph routing (KGGR) framework that captures the interplay of features from different semantic regions to help learn more powerful contextualized image representation and guides the propagation of semantic information through different categories to help train the classifiers.

The framework builds on two graph propagation networks that perform message propagation on feature and semantic spaces. Specifically, it first constructs a graph based on statistical label co-occurrences to correlate different categories. For the feature space, we designed a semantically guided attention module that applies category semantics to guide learning category-related image features that focus more on the corresponding semantic regions. By initializing the graph node with the feature vector of the corresponding category, we introduce a graph propagation network to propagate features through the graph to capture feature interactions and learn contextualized feature representations. Regarding the semantic space, we treat each node as the classifier weight of each category and utilize another graph propagation network to transfer node messages throughout the graph, allowing each category to derive information from its correlated categories to improve classifier training.

A preliminary version of this work was presented as a conference paper \cite{chen2019learning}. In this version, we inherit the idea of explicitly integrating statistical category correlations with deep graph propagation networks and strengthen the framework from several aspects as follows. First, we propagate information in both the feature and semantic spaces, which guides the learning of more powerful contextualized features and simultaneously regularizes learning classifier weights. Second, we generalize the proposed framework to the more challenging multi-label few-shot learning task and demonstrate its superiority on this task. Finally, we conduct more extensive experiments and analyses on several widely used benchmarks and demonstrate the effectiveness of the proposed framework while verifying the contribution of each component.

In summary, the contributions of this work can be summarized as follows: 1) We propose a novel knowledge-guided graph routing (KGGR) framework that explores category interplay in both feature and semantic spaces under the explicit guidance of statistical category co-occurrence correlations. This approach can help learn more powerful contextualized features and simultaneously regularize learning classifier weights. 2) We propose a simple yet effective semantically guided attention mechanism that exploits category semantics to learn semantic-aware features that focus more on semantic regions of corresponding categories. 3) We apply the KGGR framework to address both multi-label image recognition and multi-label few-shot learning tasks and conduct experiments on various benchmarks, including PASCAL VOC 2007 \& 2012 \cite{everingham2010pascal}, Microsoft-COCO \cite{lin2014microsoft}, and Visual Genome with larger-scale categories \cite{krishna2017visual}, and demonstrate that our framework exhibits substantial performance improvements on both tasks.

\section{Related Works}

In this section, we review the related works following 
three 
main research streams: multi-label image recognition, few-shot learning, and knowledge representation learning.

\subsection{Multi-label Image Recognition}
Recent progress on multi-label image classification relies on a combination of object localization and deep learning techniques \cite{wei2016hcp,yang2016exploit}. Generally, such works introduced object proposals \cite{zitnick2014edge} that were assumed to include all possible foreground objects in the image; then, they aggregated the features extracted from all these proposals to incorporate local information. Although these methods have achieved notable performance improvement, the region candidate localization step usually incurred redundant computation costs and prevented the models from being applied to end-to-end training approaches with deep neural networks. Zhang et al. \cite{zhang2018multi} further utilized a learning-based region proposal network and integrated it with deep neural networks. Although this method could be jointly optimized, it required additional bounding box annotations to train the proposal generation component. To solve this issue, some other works \cite{zhu2017learning,wang2017multi} resorted to attention mechanisms to locate the informative regions. These methods could be trained with image-level annotations in an end-to-end manner. For example, Wang et al. \cite{wang2017multi} introduced a spatial transformer to adaptively search for semantic regions and then aggregated the features from these regions to identify multiple labels. However, due to a lack of supervision and guidance, these methods could merely locate the regions roughly.

Modeling label dependencies can help capture label co-occurrence, which is also a key aspect of multi-label recognition. To achieve this, a series of works have introduced graphic models such as the conditional random field \cite{ghamrawi2005collective}, dependency network \cite{guo2011multi}, or co-occurrence matrix \cite{xue2011correlative} to capture pairwise label correlations. Recently, Wang et al. \cite{wang2016cnn} formulated a CNN-RNN framework that implicitly utilized semantic redundancy and co-occurrence dependency to facilitate effective multi-label classification. Some works \cite{zhang2018multi,chen2018recurrent} further capitalized on the proposal generation/visual attention mechanism to search local discriminative regions and used RNN/LSTM \cite{hochreiter1997long} models to explicitly model label dependencies. For example, Chen et al. \cite{chen2018recurrent} proposed a recurrent attention reinforcement learning framework to iteratively discover a sequence of attentional and informative regions to model long-term dependencies among them to help capture semantic label co-occurrences. Furthermore, Chen et al. \cite{chen2019multi} directly mapped the GloVe vector to the corresponding object classifier via a graph convolutional network. Different from all these methods, we correlate all label pairs in the form of a structured graph and introduce two parallel graph neural networks to simultaneously explore interactions among category-specific features to learn contextualized feature representation and to explore information propagation from correlated categories to learn classifier weights. Besides, the semantically guided attention mechanism that introduces category semantic to guide learning category-specific features is also original to current works.



\subsection{Few-shot learning}
Few-shot learning aims to understand novel concepts from only a few examples, mimicking the human cognitive abilities. Learning robust feature representations and training classifiers to recognize novel categories is the essential image recognition issue in the few-shot scenario. To address this task, the existing works are mainly characterized by three types of helpful techniques: metric learning, meta-learning and data synthesis, which are often combined. Metric learning-based methods \cite{vinyals2016matching,kim2019variational,snell2017prototypical,sung2018learning} are intended to learn an appropriate embedding space in which same-category samples are close while different-category samples are distant. A matching network using an attention mechanism and memory unit was introduced in \cite{vinyals2016matching} to compare the test and support samples. Snell et al. \cite{snell2017prototypical} proposed a prototypical network that adopts the mean of the embedded sample features in novel categories as a class prototype and then recognizes test samples using nearest-neighbor classifiers. In \cite{sung2018learning}, a relational network was proposed to learn a transferable deep metric. Some recent works have further adaptively generated the classifier weights of novel categories from feature embedding \cite{qiao2018few,qi2018low}. Meta-learning, also known as learning to learn, seeks to transfer some ``meta knowledge'' from previously learned tasks to achieve rapid learning of new tasks. Meta-learning is popular among few-shot learning approaches \cite{sung2018learning,Ravichandran2019ICCV,finn2017model,ravi2016optimization}. Finn et al. \cite{finn2017model} proposed a model-agnostic meta-learning approach to initialize the network well and make fine-tuning the model easier for novel categories with few samples. In \cite{ravi2016optimization}, transferable optimization strategies are learned using an LSTM-based meta-learner to achieve effective model training in the few-shot regime. Data synthesis tries to generate new samples from the few given training samples to augment model training in the few-shot task, motivated by the human ability to generate new object examples from known primitives. In addition to simple data augmentation tricks such as horizontal flipping, scaling and positional shifts, more sophisticated approaches have been introduced \cite{hariharan2017low,SchwartzKSHMKFG18nips,Tokmakov2019ICCV,AlfassyKASHFGB19cvpr}. In \cite{SchwartzKSHMKFG18nips}, a delta-encoder model was designed that learned to extract transferable intraclass deformations and synthesize samples for novel categories. Alfassy et al. \cite{AlfassyKASHFGB19cvpr} proposed a label-set operations network to synthesize samples with multiple labels to address the new and challenging multi-label few-shot task 
and was the first to introduce a benchmark. 
In this work, we generalize our KGGR framework to this task and demonstrate its superior ability to learn robust features and classifiers from limited samples.

\subsection{Knowledge Representation Learning}

Deep neural networks have made impressive breakthroughs in recent years and are capable of learning powerful representations from raw training data \cite{he2016deep,ren2017faster,chen2016disc,chen2018learning}. However, these models also have a bottleneck. One potential reason is that they lack the notable human capability of exploiting prior knowledge, which arises from knowledge accumulation and reasoning. Thus, the ability to incorporate prior knowledge into deep representation learning has become a new trend, and promising progress has been made in computer vision for tasks ranging from object detection/recognition \cite{marino2017more,jiang2018hybrid,TangWWGDGC18pami,chen2018fine,xie2020adversarial} to visual navigation/reasoning \cite{chen2019knowledge,wang2018deep,chen2018neural,WeiYang2019ICLR}. In \cite{marino2017more}, Marino et al. constructed a knowledge graph to correlate category relationships and introduced a graph search neural network to incorporate the knowledge graph into end-to-end representation learning for image classification. Tang et al. \cite{TangWWGDGC18pami} incorporated object similarity knowledge from the visual and semantic domains during the transfer learning process to help improve semisupervised object detection. To address the zero-shot recognition task, Wang et al. \cite{wang2018zero} proposed building on a graph convolutional network to utilize both semantic embeddings and categorical relationships for classifier prediction. A hierarchical semantic embedding framework was presented in \cite{chen2018fine} to incorporate hierarchal category knowledge into a deep neural network to enhance fine-grained representation learning and recognition. In \cite{wang2018deep}, an end-to-end trainable graph reasoning model was introduced to explicitly model the interplay of knowledge of people and contextual objects to perform reasoning regarding social relationships. Li et al. \cite{li2019large} introduced hierarchical category knowledge via semantic clustering \cite{yang2020fast} and used it to regularize the learning of transferable visual features for large-scale few-shot image classification. Yang et al. \cite{WeiYang2019ICLR} studied incorporating semantic priors to address the semantic navigation task, in which the prior knowledge was fused into a deep reinforcement learning framework by graph convolutional networks. In \cite{Zhang2020AAAI}, a knowledge integration network was designed to encode human and scene knowledge for action recognition.
Inspired by these works, we propose a new knowledge-guided graph routing (KGGR) framework that incorporates prior knowledge to propagate both feature and semantic space information to effectively boost multi-label analysis performance.

\begin{figure*}[!t]
\centering
\includegraphics[width=1.0\linewidth]{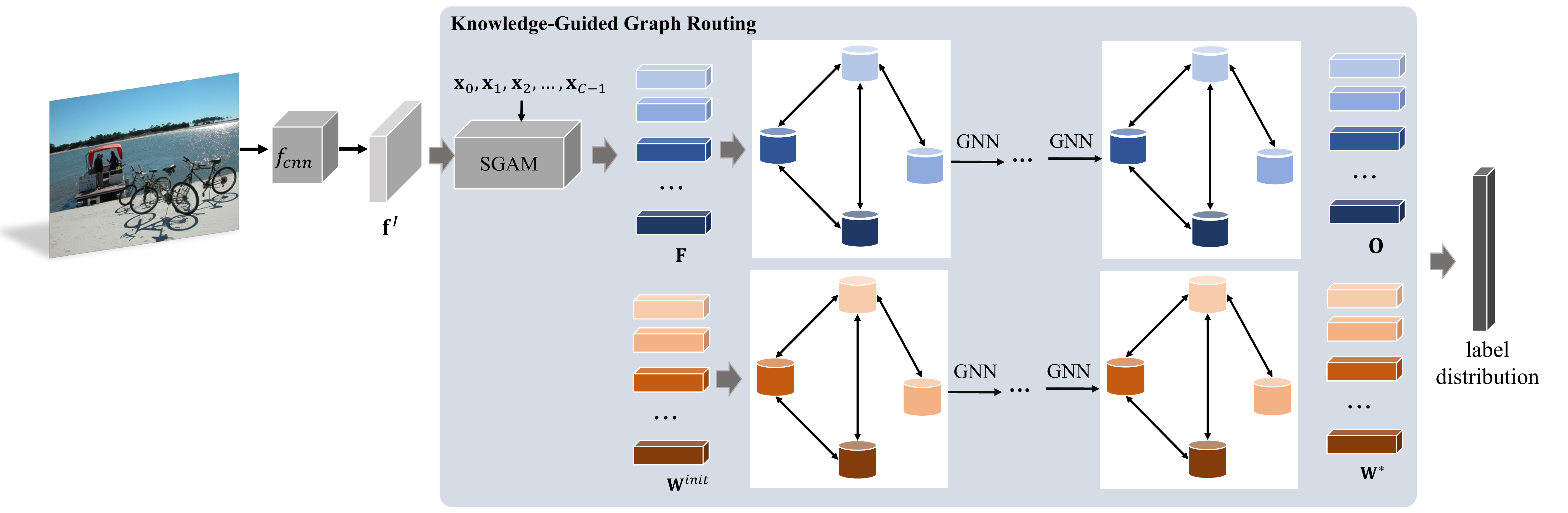} 
\caption{Illustration of our knowledge-guided graph routing framework. It first uses a CNN to extract image features, and then introduces two graph propagation networks to transfer message on both feature and semantic spaces to learn contextualized features and simultaneously regularize learning classifier weights.}
\label{fig:pipeline}
\end{figure*}

\section{KGGR Framework}
In this section, we provide an overall introduction to the proposed KGGR framework. The framework builds on a graph that encodes prior knowledge of category correlations. Then, it introduces two graph propagation networks that propagate information in both the feature and semantic spaces. For feature propagation, the framework first feeds the input image into a fully convolutional network to generate the feature maps and incorporates category semantics to guide learning semantic-specific feature vectors that focus more on the semantic regions for each category. Then, it further initializes the graph nodes with the corresponding feature vectors and applies a graph propagation network to explore feature interplay to learn contextualized features. For semantic propagation, each node is initialized with the classifier weights of the corresponding category. Then, another graph propagation network is learned to adaptively propagate node messages through the graph to explore node interactions and transfer the classifier's information of the correlated categories to help train the classifiers. The overall pipeline of the proposed KGGR framework is presented in Figure \ref{fig:pipeline}.

\subsection{Knowledge Graph Construction}
\label{sec:kgc}
We first introduce the knowledge graph $\mathcal{G}=\{\mathbf{V}, \mathbf{A}\}$, in which the nodes represent categories and the edges represent co-occurrences between corresponding categories. Specifically, suppose that the dataset covers $C$ categories, $\mathbf{V}$ can be represented as $\{v_0, v_2, \dots, v_{C-1}\}$ where element $v_c$ denotes category $c$ and $\mathbf{A}$ can be represented as $\{a_{00}, a_{01}, \dots, a_{0(C-1)}, \dots, a_{(C-1)(C-1)}\}$ where element $a_{cc'}$ denotes the probability that an object belongs to category $c'$ when in the presence of an object belonging to category $c$. We compute the probabilities between all the category pairs using the annotations from the samples in the training set; thus, we do not introduce any additional annotation.

\subsection{Graph Feature Propagation}
Here, we describe the feature propagation module, which first decouples the image into $C$ category-specific feature vectors via a semantically guided attention mechanism and then propagates these vectors throughout the graph to learn contextualized features for each category via a knowledge-embedded propagation mechanism.

\subsubsection{Semantically guided attention mechanism}
In this part, we introduce the semantically guided attention mechanism, which adopts the category semantic to guide the learning of semantic-specific feature representations. Specifically, given an input image $I$, this mechanism first extracts feature maps $\mathbf{f}^I \in \mathcal{R}^{W\times H \times N}$, where $W$, $H$, and $N$ are the width, height and channel numbers of the feature maps, respectively, formulated as
\begin{equation}
\mathbf{f}^I=f_{cnn}(I),
\label{eq:image-feature-extraction}
\end{equation}
where $f_{cnn}(\cdot)$ is a feature extractor implemented by a fully convolutional network. For each category $c$, the framework extracts a $d_s$-dimensional semantic-embedding vector using the pretrained GloVe \cite{pennington2014glove} model
\begin{equation}
\mathbf{x}_c=f_{g}(w_c),
\label{eq:semantic-feature-extraction}
\end{equation}
where $w_c$ is the semantic word representing category $c$. Then, we incorporate the semantic vector $\mathbf{x}_c$ to guide the focus to the semantic-aware regions and thus learn a feature vector corresponding to this category. More specifically, for each location $(w, h)$, we first fuse the corresponding image features $\mathbf{f}^I_{wh}$ and $\mathbf{x}_c$ using a low-rank bilinear pooling method \cite{kim2016hadamard}
\begin{equation}
\tilde{\mathbf{f}}^I_{c,wh}=\mathbf{P}^T\left(\tanh\left((\mathbf{U}^T\mathbf{f}^I_{wh})\odot(\mathbf{V}^T\mathbf{x}_c)\right)\right)+\mathbf{b},
\end{equation}
where $\tanh(\cdot)$ is the hyperbolic tangent function, and $\mathbf{U}\in \mathcal{R}^{N\times d_1}$, $\mathbf{V}\in \mathcal{R}^{d_s\times d_1}$, $\mathbf{P}\in \mathcal{R}^{d_1\times d_2}$, and $\mathbf{b}\in \mathcal{R}^{d_2}$ are the learnable parameters, $\odot$ represents elementwise multiplication, and $d_1$ and $d_2$ are the dimensions of the joint embeddings and the output features, respectively. Then, we compute an attentional coefficient under the guidance of $\mathbf{x}_c$ as follows:
\begin{equation}
\tilde{a}_{c,wh}=f_{a}(\tilde{\mathbf{f}}^I_{c,wh}),
\end{equation}
which indicates the importance of location $(w, h)$. $f_{a}(\cdot)$ is an attentional function implemented by a fully connected network. The process is repeated for all locations. To allow the coefficients to be easily compared across different samples, we normalize them over all the locations using a softmax function:
\begin{equation}
a_{c,wh}=\frac{\mathrm{exp}(\tilde{a}_{c,wh})}{\sum_{w',h'}{\mathrm{exp}(\tilde{a}_{c,w'h'}})}.
\end{equation}
Finally, we perform weighted average pooling over all the locations to obtain a feature vector
\begin{equation}
\mathbf{f}_{c}=\sum_{w,h}a_{c,wh}\mathbf{f}_{c,wh}
\end{equation}
that encodes the information related to category $c$. We repeat the process for all categories and obtain all the category-related feature vectors $\{\mathbf{f}_{0}, \mathbf{f}_{1}, \dots, \mathbf{f}_{C-1}\}$. In this way, we can learn the feature $\mathbf{f}_c$, which focuses more on the semantic regions of each category $c$ and primarily encodes the features for each category. We also present some examples for quantitative analysis in Fig. \ref{fig:semantic-map}.

\subsubsection{Knowledge-embedded feature propagation}
\label{sec:SI}
After obtaining the feature vectors corresponding to all the categories, we initialize the nodes of $\mathcal{G}$ with the feature vector of the corresponding category and introduce a graph neural network to propagate messages through the graph to explore their interactions and learn a contextualized representation.

Inspired by the gated graph neural networks method \cite{li2016gated}, we adopt a gated recurrent update mechanism to propagate messages through the graph and learn the contextualized node-level features. Specifically, each node $v_c \in \mathbf{V}$ has a hidden state $\mathbf{h}_c^t$ at timestep $t$. In this work, because each node corresponds to a specific category and our model aims to explore the interactions among semantic-specific features, we initialize the hidden state at $t=0$ with the feature vector that relates to the corresponding category, formulated as
\begin{equation}
\mathbf{h}_c^0=\mathbf{f}_{c}.
\end{equation}
At timestep $t$, the framework aggregates messages from its neighbor nodes as follows:
\begin{equation}
\mathbf{a}_c^t=\left[\sum_{c'}(a_{cc'})\mathbf{h}_c^{t-1}, \sum_{c'}(a_{c'c})\mathbf{h}_c^{t-1}\right].
\end{equation}
In this way, the framework encourages message propagation when node $c'$ is highly correlated with node $c$; otherwise, it suppresses propagation. Therefore, this approach propagates messages through the graph and explores node interactions under the prior knowledge guidance of statistical label co-occurrences. Then, the framework updates the hidden state based on the aggregated feature vector $\mathbf{a}_c^t$ and its hidden state at the previous timestep $\mathbf{h}_c^{t-1}$ via a gated mechanism similar to the gated recurrent unit, formulated as
\begin{equation}
\begin{split}
\mathbf{z}_c^t=&{}\sigma(\mathbf{W}^z{\mathbf{a}_c^t}+\mathbf{U}^z{\mathbf{h}_c^{t-1}}) \\
\mathbf{r}_c^t=&{}\sigma(\mathbf{W}^r{\mathbf{a}_c^t}+\mathbf{U}^r{\mathbf{h}_c^{t-1}}) \\
\widetilde{\mathbf{h}_c^t}=&{}\tanh\left(\mathbf{W}{\mathbf{a}_c^t}+\mathbf{U}({\mathbf{r}_c^t}\odot{\mathbf{h}_c^{t-1}})\right) \\
\mathbf{h}_c^t=&{}(1-{\mathbf{z}_c^t}) \odot{\mathbf{h}_c^{t-1}}+{\mathbf{z}_c^t}\odot{\widetilde{\mathbf{h}_c^t}},
\end{split}
\label{eq:ggnn}
\end{equation}
where $\sigma(\cdot)$ is the logistic sigmoid function, $\tanh(\cdot)$ is the hyperbolic tangent function, and $\odot$ is elementwise multiplication. In this way, each node aggregates messages from the other nodes and simultaneously propagates their information through the graph, enabling interactions among all the feature vectors corresponding to all categories. The process is repeated $T_f$ times; the generated final hidden states are $\{\mathbf{h}_0^{T_f}, \mathbf{h}_1^{T_f}, \dots, \mathbf{h}_{C-1}^{T_f}\}$. Here, the hidden state of each node $\mathbf{h}_{c}^{T_f}$ not only encodes features from category $c$ but also carries contextualized messages from other categories. Finally, we concatenate $\mathbf{h}_{c}^{T_f}$ and the input feature vector $\mathbf{h}_{c}^0$ to obtain the feature vector for each category
\begin{equation}
\begin{split}
\mathbf{o}_c=f_o(\mathbf{h}_{c}^{T_f}, \mathbf{h}_{c}^0) \\
\end{split}
\label{eq:cls}
\end{equation}
where $f_o(\cdot)$ is an output function that maps the concatenation of $\mathbf{h}_{c}^T$ and $\mathbf{h}_{c}^0$ into an output vector $\mathbf{o}_c$.

\subsection{Graph Semantic Propagation}
A classifier weight can be regarded as the prototype representation of the corresponding category. Transferring the prototype representations among similar categories can substantially enhance classifier training, especially for categories with few training samples. To achieve this, we integrate the graph $\mathcal{G}$ to guide the prototype representation propagation among all the categories to fully exploit the information of correlated categories and improve classifier training. Specifically, we initialize each graph node with the classifier weight of the corresponding category and iteratively propagate the node messages though the graph. In each iteration $t$, node $v_c$ has a hidden state $\mathbf{h}_c^t$. The hidden state $\mathbf{h}_c^0$ at iteration $t=0$ is set to the initial classifier weight $\mathbf{w}_{c}^{init}$, formulated as follows:
\begin{equation}
\mathbf{h}_c^0=\mathbf{w}_{c}^{init},
\label{eq:node-initialization}
\end{equation}
where $\mathbf{w}_c^{init}$ is randomly initialized before training. At each iteration $t$, each node $k$ aggregates the messages from its correlated nodes such that the parameter vectors of those nodes can help refine the node's own parameter vector, formulated as follows:
\begin{equation}
\mathbf{a}_c^t=[\sum_{c'=1}^K{a_{kk'}\mathbf{h}_{c}^{t-1}},\sum_{c'=1}^K{a_{c'c}\mathbf{h}_{c}^{t-1}}].
\label{eq:node-aggregation}
\end{equation}
In this way, a high correlation between nodes $k$ and $k'$ encourages message propagation from $k'$ to $k$; otherwise, it suppresses the propagation. Then, the framework uses these aggregated feature vectors and the hidden state of the previous iteration as input to update the hidden state via a gated mechanism as Eq. \ref{eq:ggnn}. In this way, each node can aggregate more message from the nodes of correlated categories to help update its classifiers. The iteration is repeated $T_s$ times; eventually, the final hidden states $\{\mathbf{h}_0^{T_s}, \mathbf{h}_1^{T_s}, \dots, \mathbf{h}_{C-1}^{T_s}\}$ are generated. Finally, we also utilize a simple output network to predict the classifier weight
\begin{equation}
\mathbf{w}_c^*=w_o(\mathbf{h}_c^{T_s}, \mathbf{h}_c^0).
\label{eq:node-output}
\end{equation}

\subsection{Network Architecture}
Following existing multi-label image classification works \cite{zhu2017learning}, we implement the feature extractor $f_{cnn}(\cdot)$ based on the widely used ResNet-101 \cite{he2016deep}. Specifically, we replace the last average pooling layer with another average pooling layer with a size of $2\times 2$ and a stride of 2. The other layers remain unchanged in this implementation. For the low rank bilinear pooling operation, $N$, $d_s$, $d_1$, and $d_2$ are set to 2,048, 300, 1,024, and 1,024, respectively. Thus, $f_a(\cdot)$ is implemented by a 1,024-to-1 fully connected layer that maps the 1,024-feature vector to a single attentional coefficient.

The two gate graph networks share the same architecture. Specifically, the hidden state dimension is set to 2,048, while the iteration numbers $T_f$ and $T_s$ are set to 3. The output vector dimension $\mathbf{o}_c$ is also set to 2,048. Thus, the output networks $f_o(\cdot)$ and $w_o(\cdot)$ are implemented by a 4,096-to-2,048 fully connected layer followed by the hyperbolic tangent function.

\subsection{From MLR to ML-FSL}
As we obtain the feature vector $\mathbf{f}_c$ and classifier $\mathbf{w}_c^*$ for each category $c$, we multiply them to obtain the score:
\begin{equation}
s_c=\mathbf{w}_c^{*\top}\mathbf{f}_c.
\label{eq:score}
\end{equation}
We conduct this multiplication process for all the categories and obtain a score vector $\mathbf{s}=\{s_0, s_1, \dots, s_{C-1}\}$. Next, we apply the framework to multi-label image recognition and adapt it to a more challenging task, i.e., multi-label few shot learning.

\subsubsection{Multi-label Image Recognition}
\label{sec:mlr}
Given a dataset that contains $M$ training samples $\{I_i, y_i\}_{i=0}^{M-1}$, in which $I_i$ is the $i$-th image and $y_i=\{y_{i0}, y_{i1}, \dots, y_{i(C-1)}\}$ is the corresponding annotation. $y_{ic}$ is assigned a 1 if the sample is annotated with category $c$ and 0 otherwise. Given an image $I_i$, we can obtain a predicted score vector $\mathbf{s}_i=\{s_{i0}, s_{i1}, \dots, s_{i(C-1)}\}$ and compute the corresponding probability vector $\mathbf{p}_i=\{p_{i0}, p_{i1}, \dots, p_{i(C-1)}\}$ via a sigmoid function
\begin{equation}
p_{ic}=\sigma(s_{ic}).
\end{equation}
We adopt cross entropy as the objective loss function
\begin{equation}
\mathcal{L}=\sum_{i=0}^{N-1}\sum_{c=0}^{C-1}\left(y_{ic}\log p_{ic}+(1-y_{ic})\log(1-p_{ic})\right).
\label{eq:mlr_loss}
\end{equation}

We train the proposed framework with the loss $\mathcal{L}$ in an end-to-end fashion. Specifically, we first apply the ResNet-101 parameters pretrained on the ImageNet dataset \cite{deng2009imagenet} to initialize the parameters of the corresponding layers in $f_{cnn}$; we initialize the parameters of other layers randomly. Because the lower layer parameters pretrained on the ImageNet dataset generalize well across different datasets, we fix the parameters of the previous 92 convolutional layers in $f_{cnn}(\cdot)$ and jointly optimize all the other layers. The framework is trained with the Adam algorithm \cite{kingma2015adam} with a batch size of 4 and momentums of 0.999 and 0.9. The learning rate is initialized to $10^{-5}$ and divided by 10 when the error plateaus. During training, the input image is resized to $640\times 640$, and we randomly choose a number from $\{640, 576, 512, 448, 384, 320\}$ as the width and height to crop random patches. Finally, the cropped patches are further resized to $576\times 576$. During testing, we simply resize the input image to $640\times 640$ and perform a center crop with a size of $576\times 576$ for evaluation.

\subsubsection{Multi-Label Few-Shot Learning}
For multi-label few-shot learning, the dataset contains a set of $C_b$ base categories with sufficient training samples and a set of $C_n$ novel categories with limited training samples (e.g., 1, 2, or 5). Because the training samples for the base and novel categories are extremely unbalanced, it inevitably leads to poor performances if using the loss as Equation (\ref{eq:mlr_loss}) for training, especially on the novel categories. Thus, we follow the previous work \cite{alfassy2019laso} to adopt a two-stage process to train the proposed framework.

\noindent\textbf{Stage 1.} In the first stage, we train the framework using the training samples of the base categories. Here, we have $C_b$ categories, and construct the adjacent matrix $\mathbf{A}=\mathbf{A}_b\in \mathcal{R}^{C_b \times C_b}$ based on the co-occurrences among these $C_b$ categories (following the process described in Sec. \ref{sec:kgc}). Given a sample from the $C_b$ categories, we can compute the score vector $\mathbf{s}_i=\{s_{i0}, s_{i1}, \dots, s_{i(C_b-1)}\}$ and the corresponding probability vector $\mathbf{p}_i=\{p_{i0}, p_{i1}, \dots, p_{i(C_b-1)}\}$. Then, we define the loss similarly to Eq. \ref{eq:mlr_loss}, expressed as 
\begin{equation}
\mathcal{L}_b=\sum_{i=0}^{N_b-1}\sum_{c=0}^{C_b-1}\left(y_{ic}\log p_{ic}+(1-y_{ic})\log(1-p_{ic})\right).
\label{eq:mlr_fsl_loss}
\end{equation}
where $N_b$ is the number of training samples from the base set. Similarly, the parameters of the backbone network are initialized with the parameters pretrained on the ImageNet dataset, while those of the other layers are randomly initialized. We also fix the previous convolutional layers in the backbone networks and jointly train all the other layers using the Adam algorithm with a batch size of 4 and momentums of 0.999 and 0.9. We set the initial learning rate to $10^{-5}$ and divided it by 10 after 12 epochs. The framework is trained with 20 epochs in total. During training, the augmentation strategy is the same as that adopted for multi-label recognition (described in Sec. \ref{sec:mlr}).

\noindent\textbf{Stage 2.} In the second stage, we fix the parameters of the backbone network and the attention modules for feature extraction and train the two graph neural networks using the novel sets. Because the training samples are limited during this stage, we cannot compute the statistical co-occurrence to obtain the adjacent matrix. To address this issue, we consider semantic similarity to compute the label correlations. Specifically, we use Glove \cite{pennington2014glove} to extract semantic vector for each category, compute the cosine distance between each category pair to obtain their semantic similarities, and construct the adjacent matrix $\mathbf{A}=\mathbf{A}_n\in \mathcal{R}^{C_n \times C_n}$. In addition, we introduce an additional regularization term on the classifier weights and thus the loss function can be formulated as
\begin{equation}
\begin{split}
\mathcal{L}_b=&\sum_{i=0}^{N_n-1}\sum_{c=0}^{C_n-1}\left(y_{ic}\log p_{ic}+(1-y_{ic})\log(1-p_{ic})\right) \\
&+\gamma\sum_{c=0}^{C_n-1}||\mathbf{w}^*_c||^2_2,
\end{split}
\end{equation}
where $N_n$ is the number of samples from the novel set and $\gamma$ is a balance parameter that is set to 0.001. We train the model using the Adam algorithm with the same batch size, momentums, and data augmentation strategy as in stage 1. Because we merely train the two graph neural networks while fixing other layers, we set a large learning rate $10^{-4}$ and train the framework for merely 500 iterations.

\begin{table*}[!t]
\centering
\begin{tabular}{c|c|cccccc|cccccc}
\hline
& & \multicolumn{6}{c|}{Top 3} &\multicolumn{6}{c}{All} \\
\hline
\centering Methods & mAP & CP & CR & CF1 & OP & OR & OF1 & CP & CR & CF1 & OP & OR & OF1 \\
\hline
\hline
WARP \cite{gong2014deep} &- & 59.3 & 52.5 & 55.7 & 59.8 & 61.4 & 60.7 & - & - & -&- & -&- \\
CNN-RNN \cite{wang2016cnn} &- & 66.0 & 55.6 & 60.4 & 69.2 & 66.4 & 67.8 &- & - & -&- & -&-\\
RLSD \cite{zhang2018multi} &- & 67.6 & 57.2 & 62.0 & 70.1 & 63.4 & 66.5 &- & - & -&- & -&-\\
RARL \cite{chen2018recurrent} & - & 78.8 & 57.2 & 66.2 & 84.0 & 61.6 & 71.1 & \\
RDAR \cite{wang2017multi} & 73.4 & 79.1 & 58.7 & 67.4 & 84.0 & 63.0 & 72.0 &- & - & -&- & -&-\\
KD-WSD \cite{DBLP:conf/mm/LiuSSYXP18}& 74.6 & - & -& 66.8 & - &- &72.7 &- &-& 69.2 & - & -&74.0 \\
ResNet-SRN-att \cite{zhu2017learning} & 76.1 & 85.8 & 57.5 & 66.3 & 88.1 & 61.1 & 72.1 & 81.2 & 63.3 & 70.0 & 84.1 & 67.7 & 75.0\\
ResNet-SRN \cite{zhu2017learning} & 77.1 & 85.2 & 58.8 & 67.4 & 87.4 & 62.5 & 72.9 & 81.6 & 65.4 & 71.2 & 82.7 & 69.9 & 75.8 \\
ML-GCN \cite{DBLP:conf/cvpr/ChenWWG19} & 83.1 & 89.2 & 64.1 & 74.6 & 90.5 & 66.5 & 76.7 &85.1 & 72.0 & 78.0 & 85.8 &75.4 & 80.3 \\
\hline
Ours & \textbf{84.3} & \textbf{89.4} & \textbf{64.6} & \textbf{75.0} & \textbf{91.3} & \textbf{66.6} & \textbf{77.0} & \textbf{85.6} & \textbf{72.7} & \textbf{78.6} & \textbf{87.1} & \textbf{75.6} & \textbf{80.9} \\
\hline
\end{tabular}
\caption{Comparison of the mAP, CP, CR, CF1 and OP, OR, OF1 (in \%) scores of our framework and other state-of-the-art methods under settings of all and top-3 labels on the Microsoft COCO dataset. A hyphen ``-'' denotes that no corresponding result was provided.}
\vspace{-10pt}
\label{table:result-coco}
\end{table*}

\section{Experiments}
In this section, we conduct extensive experiments on various benchmarks to evaluate the performance of the proposed framework over existing state-of-the-art methods and perform ablative studies to further analyze the actual contribution of each component.

\subsection{Evaluation Metrics}
For multi-label image recognition, we follow current state-of-the-art works \cite{wei2016hcp,chen2019multi} by adopting the average precision (AP) on each category and the mean average precision (mAP) over all categories. For more comprehensive comparisons, we further follow the current works \cite{zhu2017learning,li2017improving} to present the precision, recall, and F1-measure values. Here, we assign the labels with the top-$3$ highest scores for each image and compare them with the ground truth labels. Concretely, we adopt overall precision, recall, and F1-measure (OP, OR, and OF1) and per-class precision, recall, and F1-measure (CP, CR, and CF1), which are defined as shown below:
\begin{equation}
\begin{split}
\mathrm{OP}&=\frac{\sum_{i}N_{i}^{c}}{\sum_{i}N_{i}^{p}},\quad\\
\mathrm{OR}&=\frac{\sum_{i}N_{i}^{c}}{\sum_{i}N_{i}^{g}},\quad\\
\mathrm{OF}1&=\frac{2 \times \mathrm{OP} \times \mathrm{OR}}{\mathrm{OP}+\mathrm{OR}},\quad
\end{split}
\begin{split}
\mathrm{CP}&=\frac{1}{C}\sum_{i}\frac{N_{i}^{c}}{N_{i}^{p}},\\
\mathrm{CR}&=\frac{1}{C}\sum_{i}\frac{N_{i}^{c}}{N_{i}^{g}},\\
\mathrm{CF}1&=\frac{2 \times \mathrm{CP} \times \mathrm{CR}}{\mathrm{CP}+\mathrm{CR}},
\end{split}
\label{eqn:metric}
\end{equation}
where $C$ is the number of labels, $N_{i}^{c}$ is the number of images for which the $i$-th label is correctly predicted, $N_{i}^{p}$ is the number of images for which the $i$-th label is predicted, and $N_{i}^{g}$ is the number of ground truth images for the $i$-th label. The above metrics require a fixed number of labels, but the label numbers of different images generally vary. Thus, we further present the OP, OR, OF1 and the CP, CR, CF1 metrics using the idea that a label is predicted as positive when its estimated probability is greater than 0.5 \cite{zhu2017learning}. Among these metrics, mAP, OF1, and CF1 are the most important and provide a more comprehensive evaluation.


\subsection{Datasets}

\noindent\textbf{Pascal VOC 2007 \& 2012} \cite{everingham2010pascal} are the datasets most widely used to evaluate the multi-label image classification task, and most of the existing works report their results on these datasets. Therefore, we also conducted experiments on these two datasets for evaluation purposes. Specifically, both datasets include 20 common categories. Pascal VOC 2007 contains a training and validation (trainval) set with 5,011 images and a test set with 4,952 images, while VOC 2012 includes 11,540 images as the trainval set and 10,991 images as the test set. For fair comparison purposes, the proposed framework and existing competitors were all trained on the trainval set and evaluated on the test set. However, because the number of categories in these two datasets is limited, we use these datasets merely to evaluate the multi-label image recognition task; we do not use them to evaluate the multi-label few-shot learning task.

\noindent\textbf{Microsoft COCO} \cite{lin2014microsoft} was originally constructed for object detection and segmentation but has recently been adopted to evaluate multi-label image classification. This dataset contains 122,218 images and covers 80 common categories. The dataset is further divided into a training set with 82,081 images and a validation set with 40,137 images. Because no ground-truth annotations are available for the test set, we trained our method and all the compared methods on the training set and evaluated them on the validation set.

For multi-label few-shot learning, we follow the previous work \cite{alfassy2019laso} to split the 80 categories into 64 base categories and 16 novel categories. Specifically, the novel categories are \emph{bicycle, boat, stop sign, bird, backpack, frisbee, snowboard, surfboard, cup, fork, spoon, broccoli, chair, keyboard, microwave,} and \emph{vase}. To ensure fair comparisons with current works, we utilize the trainval samples of the base categories as the base set and randomly select $K$ ($K$=1,5) trainval samples from the novel categories as the novel set. We evaluate the test samples of the novel categories.

\noindent\textbf{Visual Genome} \cite{krishna2017visual} is a dataset that contains 108,249 images and covers 80,138 categories. Because most categories have very few samples, we consider only the 500 most frequent categories, resulting in a Visual Genome 500 (VG-500) subset. We randomly selected 10,000 images as the test set and the remaining 98,249 images as the training set. Compared with other existing benchmarks, this one covers more categories, i.e., 500 categories as opposed to 20 in Pascal VOC \cite{everingham2010pascal} and 80 in Microsoft-COCO \cite{lin2014microsoft}; thus, it can be used to evaluate multi-label image recognition performance with larger-scale categories.

For multi-label few-shot learning, we split VG-500 into 400 base categories and 100 novel categories. Similar to Microsoft COCO, we used all training samples of the base categories as the base set and selected $K$ ($K$=1,5) training samples of the novel categories as the novel set. Then, we evaluate the models on the test samples of the novel categories.

\begin{table*}[htp]
\centering
\scriptsize
\begin{tabular}
{p{2.5cm}|p{0.2cm}p{0.2cm}p{0.2cm}p{0.2cm}p{0.3cm}p{0.3cm}p{0.20cm}p{0.20cm}p{0.3cm}p{0.3cm}p{0.3cm}p{0.3cm}p{0.3cm}p{0.4cm}p{0.3cm}p{0.3cm}p{0.3cm}p{0.3cm}p{0.3cm}p{0.3cm}|p{0.5cm}}
\hline
\centering Methods & aero & bike & bird & boat & bottle & bus & car & cat & chair & cow & table & dog & horse & mbike & person & plant & sheep & sofa & train & tv & mAP \\
\hline
\hline
\centering CNN-RNN~\cite{wang2016cnn} & 96.7 & 83.1 & 94.2 & 92.8 & 61.2 & 82.1 & 89.1 & 94.2 & 64.2 & 83.6 & 70.0 & 92.4 & 91.7 & 84.2 & 93.7 & 59.8 & 93.2 & 75.3 & 99.7 & 78.6 & 84.0 \\
\centering RMIC \cite{DBLP:conf/aaai/He0G0T18} & 97.1 & 91.3 & 94.2 & 57.1 & 86.7 & 90.7 & 93.1 & 63.3 & 83.3 & 76.4 & 92.8 & 94.4 & 91.6 & 95.1 & 92.3 & 59.7 & 86.0 & 69.5 & 96.4 & 79.0 & 84.5 \\
\centering VGG16+SVM~\cite{simonyan2015very} & - & -&- & -&- & - & -&- &- & -& -& -& - & -& -&- &- & - &- &- & 89.3\\
\centering VGG19+SVM~\cite{simonyan2015very} & - & -&- & -&- & - & -&- &- & -& -& -& - & -& -&- &- & - &- &- & 89.3\\
\centering RLSD \cite{zhang2018multi} & 96.4 & 92.7 & 93.8 & 94.1 & 71.2 & 92.5 & 94.2 & 95.7 & 74.3 & 90.0 & 74.2 & 95.4 & 96.2 & 92.1 & 97.9 & 66.9 & 93.5 & 73.7 & 97.5 & 87.6 & 88.5\\
\centering HCP~\cite{wei2016hcp} & 98.6 & 97.1 & \textcolor[rgb]{0,0,1}{98.0} & 95.6 & 75.3&94.7 & 95.8 &97.3 & 73.1 & 90.2 & 80.0 & 97.3 & 96.1 & 94.9 & 96.3 & 78.3 & 94.7 & 76.2 & 97.9 & 91.5 & 90.9\\
\centering FeV+LV~\cite{yang2016exploit} & 97.9 & 97.0 & 96.6 & 94.6 & 73.6 & 93.9 & 96.5& 95.5 & 73.7 & 90.3 & 82.8 & 95.4 & 97.7 & \textcolor[rgb]{0,0,1}{95.9} & 98.6 & 77.6 & 88.7 & 78.0 & 98.3 & 89.0 & 90.6\\
\centering RDAR \cite{wang2017multi} & 98.6 & 97.4 & 96.3 & 96.2 & 75.2 & 92.4 & 96.5 & 97.1 & 76.5 & 92.0 & \textcolor[rgb]{1,0,0}{87.7} & 96.8 & 97.5 & 93.8 & 98.5 & 81.6 & 93.7 & \textcolor[rgb]{0,0,1}{82.8} & \textcolor[rgb]{0,0,1}{98.6} &89.3 & 91.9 \\
\centering RARL \cite{chen2018recurrent} & 98.6 & 97.1 & 97.1 & 95.5 & 75.6 & 92.8 & \textcolor[rgb]{0,0,1}{96.8} & 97.3 & \textcolor[rgb]{0,0,1}{78.3} & 92.2 & \textcolor[rgb]{0,0,1}{87.6} & 96.9 & 96.5 & 93.6 & 98.5 & 81.6 & 93.1 & 83.2 & 98.5 & 89.3 & 92.0 \\
\centering RCP \cite{wang2016beyond} & \textcolor[rgb]{0,0,1}{99.3} & \textcolor[rgb]{0,0,1}{97.6} & \textcolor[rgb]{0,0,1}{98.0} & 96.4 & 79.3 & 93.8 & 96.6 & 97.1 & 78.0 & 88.7 & 87.1 & 97.1 & 96.3 & 95.4 & \textcolor[rgb]{1,0,0}{99.1} & 82.1 & 93.6 & 82.2 & 98.4 & \textcolor[rgb]{0,0,1}{92.8} & 92.5 \\
\centering \textbf{Ours} & \textcolor[rgb]{1,0,0}{99.8} & 97.1 & \textcolor[rgb]{1,0,0}{98.4} & \textcolor[rgb]{0,0,1}{98.0} & \textcolor[rgb]{0,0,1}{84.2} & \textcolor[rgb]{0,0,1}{95.1} & \textcolor[rgb]{0,0,1}{96.9} & \textcolor[rgb]{0,0,1}{98.4} & \textcolor[rgb]{0,0,1}{78.6} & \textcolor[rgb]{0,0,1}{94.9} & 87.0& \textcolor[rgb]{0,0,1}{98.1} &\textcolor[rgb]{0,0,1}{97.7} & \textcolor[rgb]{1,0,0}{97.4} & \textcolor[rgb]{0,0,1}{98.7} & \textcolor[rgb]{0,0,1}{82.4} &\textcolor[rgb]{0,0,1}{97.1} & 82.5 & \textcolor[rgb]{0,0,1}{98.7} & 92.0 & \textcolor[rgb]{0,0,1}{93.6}\\
\centering \textbf{Ours (pre)} & \textcolor[rgb]{0,0,1}{99.3} & \textcolor[rgb]{1,0,0}{98.6} & 97.9 & \textcolor[rgb]{1,0,0}{98.4} & \textcolor[rgb]{1,0,0}{86.2} & \textcolor[rgb]{1,0,0}{97.0} & \textcolor[rgb]{1,0,0}{98.0} & \textcolor[rgb]{1,0,0}{99.2} & \textcolor[rgb]{1,0,0}{82.6} & \textcolor[rgb]{1,0,0}{98.3} & 87.5 & \textcolor[rgb]{1,0,0}{99.0} & \textcolor[rgb]{1,0,0}{98.9} & \textcolor[rgb]{1,0,0}{97.4} & \textcolor[rgb]{1,0,0}{99.1} & \textcolor[rgb]{1,0,0}{86.9} & \textcolor[rgb]{1,0,0}{98.2} & \textcolor[rgb]{1,0,0}{84.1} & \textcolor[rgb]{1,0,0}{99.0} & \textcolor[rgb]{1,0,0}{95.0} & \textcolor[rgb]{1,0,0}{95.0}\\
\hline
\hline
\centering VGG16\&19+SVM~\cite{simonyan2015very} & 98.9 & 95.0 & 96.8 & 95.4 & 69.7 & 90.4 & 93.5 & 96.0 & 74.2 & 86.6 & 87.8 & 96.0 & 96.3 & 93.1 & 97.2 & 70.0 & 92.1 & 80.3 & 98.1 & 87.0 & 89.7\\
\centering FeV+LV (fusion)~\cite{yang2016exploit} & 98.2 & 96.9 & 97.1 & 95.8 & 74.3 & 94.2 & 96.7 & 96.7 & 76.7 & 90.5 & 88.0 & 96.9 & 97.7 & \textcolor[rgb]{0,0,1}{95.9} & 98.6 & 78.5 & 93.6 & 82.4 & 98.4 & 90.4 & 92.0 \\
\hline
\end{tabular}
\vspace{1pt}
\caption{Comparison of the AP and mAP (in \%) results of our framework and those of state-of-the-art methods on the PASCAL VOC 2007 dataset. The upper part presents the results of single models and the lower part presents those that aggregate multiple models.
``Ours'' and ``Ours (pre)'' denote our framework without and with pretraining on the COCO dataset, respectively.
The best and second-best results are highlighted in {\color{red}{red}} and {\color{blue}{blue}}, respectively. A hyphen ``-'' denotes that no corresponding result was provided. Best viewed in color.}
\label{table:comparision_voc07}
\end{table*}

\begin{table*}[htp]
\centering
\scriptsize
\begin{tabular}
{p{2.5cm}|p{0.2cm}p{0.2cm}p{0.2cm}p{0.2cm}p{0.3cm}p{0.3cm}p{0.20cm}p{0.20cm}p{0.3cm}p{0.3cm}p{0.3cm}p{0.3cm}p{0.3cm}p{0.4cm}p{0.3cm}p{0.3cm}p{0.3cm}p{0.3cm}p{0.3cm}p{0.3cm}|p{0.5cm}}
\hline
\centering Methods & aero & bike & bird & boat & bottle & bus & car & cat & chair & cow & table & dog & horse & mbike & person & plant & sheep & sofa & train & tv & mAP \\
\hline
\hline
\centering RMIC \cite{DBLP:conf/aaai/He0G0T18} & 98.0 & 85.5 & 92.6 & 88.7 & 64.0 & 86.8 & 82.0 & 94.9 & 72.7 & 83.1 & 73.4 & 95.2 & 91.7 & 90.8 &95.5 & 58.3 & 87.6 & 70.6 & 93.8 & 83.0 & 84.4 \\
\centering VGG16+SVM~\cite{simonyan2015very} & 99.0 & 88.8 & 95.9 & 93.8 & 73.1 & 92.1 & 85.1 & 97.8 & 79.5 & 91.1 & 83.3 & 97.2 & 96.3 & 94.5 & 96.9 & 63.1 & 93.4 & 75.0 & 97.1 & 87.1 & 89.0 \\
\centering VGG19+SVM~\cite{simonyan2015very} & 99.1 & 88.7 & 95.7 & 93.9 & 73.1 & 92.1 & 84.8 & 97.7 & 79.1 & 90.7 & 83.2 & 97.3 & 96.2 & 94.3 & 96.9 & 63.4 & 93.2 & 74.6 & 97.3 & 87.9 & 89.0 \\
\centering HCP \cite{wei2016hcp} & 99.1 & 92.8 & 97.4 & 94.4 & 79.9 & 93.6 & 89.8 & 98.2 & 78.2 & 94.9 & 79.8 & 97.8 & 97.0 & 93.8 & 96.4 & 74.3 & 94.7 & 71.9 & 96.7 & 88.6 & 90.5 \\
\centering FeV+LV~\cite{yang2016exploit} & 98.4 & 92.8 & 93.4 & 90.7 & 74.9 & 93.2 & 90.2 & 96.1 & 78.2 & 89.8 & 80.6 & 95.7 & 96.1 & 95.3 & 97.5 & 73.1 & 91.2 & 75.4 & 97.0 & 88.2 & 89.4 \\
\centering RCP \cite{wang2016beyond} & \textcolor[rgb]{0,0,1}{99.3} & 92.2 & \textcolor[rgb]{0,0,1}{97.5} & 94.9 & 82.3 & 94.1 & 92.4 & \textcolor[rgb]{0,0,1}{98.5} & 83.8 & 93.5 & 83.1 & 98.1 & 97.3 & 96.0 & \textcolor[rgb]{0,0,1}{98.8} & 77.7 & 95.1 & 79.4 & 97.7 & 92.4 & 92.2 \\
\centering \textbf{Ours} & \textcolor[rgb]{1,0,0}{99.6} & \textcolor[rgb]{0,0,1}{95.1} & \textcolor[rgb]{0,0,1}{97.5} & \textcolor[rgb]{1,0,0}{96.9} & \textcolor[rgb]{0,0,1}{83.1} & \textcolor[rgb]{0,0,1}{94.8} & \textcolor[rgb]{0,0,1}{94.2} & \textcolor[rgb]{0,0,1}{98.9} & \textcolor[rgb]{0,0,1}{86.5} & \textcolor[rgb]{0,0,1}{97.0} & \textcolor[rgb]{0,0,1}{84.9} & \textcolor[rgb]{0,0,1}{98.9} & \textcolor[rgb]{0,0,1}{98.9} & \textcolor[rgb]{0,0,1}{96.6} & \textcolor[rgb]{0,0,1}{98.8} & \textcolor[rgb]{0,0,1}{81.8} & \textcolor[rgb]{0,0,1}{98.3} & \textcolor[rgb]{0,0,1}{84.0} & \textcolor[rgb]{0,0,1}{98.4} & \textcolor[rgb]{0,0,1}{93.1} & \textcolor[rgb]{0,0,1}{93.9} \\
\centering \textbf{Ours (pre)} & \textcolor[rgb]{1,0,0}{99.6} & \textcolor[rgb]{1,0,0}{96.8} & \textcolor[rgb]{1,0,0}{97.9} & \textcolor[rgb]{0,0,1}{96.7} & \textcolor[rgb]{1,0,0}{87.3} & \textcolor[rgb]{1,0,0}{96.5} &  \textcolor[rgb]{1,0,0}{96.2} & \textcolor[rgb]{1,0,0}{99.1} & \textcolor[rgb]{1,0,0}{87.9} & \textcolor[rgb]{1,0,0}{97.7} & \textcolor[rgb]{1,0,0}{86.8} & \textcolor[rgb]{1,0,0}{99.3} & \textcolor[rgb]{1,0,0}{99.3} & \textcolor[rgb]{1,0,0}{97.5} & \textcolor[rgb]{1,0,0}{99.1} & \textcolor[rgb]{1,0,0}{85.39} & \textcolor[rgb]{1,0,0}{98.8} & \textcolor[rgb]{1,0,0}{84.9} &\textcolor[rgb]{1,0,0}{99.6} & \textcolor[rgb]{1,0,0}{94.4} & \textcolor[rgb]{1,0,0}{95.0} \\
\hline
\hline
\centering VGG16\&19+SVM~\cite{simonyan2015very} & \textcolor[rgb]{0,0,1}{99.1} & 89.1 & 96.0 & 94.1 & 74.1 & 92.2 & 85.3 & 97.9 & 79.9 & 92.0 & 83.7 & 97.5 & 96.5 & 94.7 & 97.1 & 63.7 & 93.6 & 75.2 & 97.4 & 87.8 & 89.3 \\
\centering FeV+LV (fusion)~\cite{yang2016exploit} & 98.9 & 93.1 & 96.0 & 94.1 & 76.4 & 93.5 & 90.8 & 97.9 & 80.2 & 92.1 &82.4 & 97.2 & 96.8 & 95.7 & 98.1 & 73.9 & 93.6 & 76.8 & 97.5 & 89.0 & 90.7 \\
\centering HCP+AGS \cite{wei2016hcp,dong2013subcategory} & \textcolor[rgb]{1,0,0}{99.8} & \textcolor[rgb]{0,0,1}{94.8} & 97.7 & 95.4 & 81.3 & 96.0 & 94.5 & 98.9 & 88.5 & 94.1 & 86.0 & 98.1 & 98.3 & 97.3 & 97.3 & 76.1 & 93.9 & 84.2 & 98.2 & 92.7 & 93.2 \\
\centering RCP+AGS \cite{wang2016beyond,dong2013subcategory} & \textcolor[rgb]{1,0,0}{99.8} & 94.5 & \textcolor[rgb]{0,0,1}{98.1} & \textcolor[rgb]{0,0,1}{96.1} & \textcolor[rgb]{0,0,1}{85.5} & \textcolor[rgb]{0,0,1}{96.1} & \textcolor[rgb]{0,0,1}{95.5} & \textcolor[rgb]{0,0,1}{99.0} & \textcolor[rgb]{1,0,0}{90.2} & \textcolor[rgb]{0,0,1}{95.0} & \textcolor[rgb]{0,0,1}{87.8} & \textcolor[rgb]{0,0,1}{98.7} & \textcolor[rgb]{0,0,1}{98.4} & \textcolor[rgb]{0,0,1}{97.5} & \textcolor[rgb]{0,0,1}{99.0} & \textcolor[rgb]{0,0,1}{80.1} & \textcolor[rgb]{0,0,1}{95.9} & \textcolor[rgb]{0,0,1}{86.5} & \textcolor[rgb]{0,0,1}{98.8} & \textcolor[rgb]{0,0,1}{94.6} & \textcolor[rgb]{0,0,1}{94.3} \\
\centering \textbf{Ours (pre \& fusion)} & \textcolor[rgb]{1,0,0}{99.8} & \textcolor[rgb]{1,0,0}{97.3} & \textcolor[rgb]{1,0,0}{98.4} & \textcolor[rgb]{1,0,0}{97.1} & \textcolor[rgb]{1,0,0}{87.9} & \textcolor[rgb]{1,0,0}{97.3} & \textcolor[rgb]{1,0,0}{96.5}  & \textcolor[rgb]{1,0,0}{99.3} & \textcolor[rgb]{0,0,1}{89.4} & \textcolor[rgb]{1,0,0}{97.8} & \textcolor[rgb]{1,0,0}{88.7} & \textcolor[rgb]{1,0,0}{99.4} & \textcolor[rgb]{1,0,0}{99.4} & \textcolor[rgb]{1,0,0}{97.9} & \textcolor[rgb]{1,0,0}{99.2} & \textcolor[rgb]{1,0,0}{86.3} & \textcolor[rgb]{1,0,0}{98.8} & \textcolor[rgb]{1,0,0}{86.3} & \textcolor[rgb]{1,0,0}{99.7} & \textcolor[rgb]{1,0,0}{95.2} & \textcolor[rgb]{1,0,0}{95.6}\\
\hline
\end{tabular}
\vspace{1pt}
\caption{Comparison of the AP and mAP scores (in \%) of our model and those of state-of-the-art methods on the PASCAL VOC 2012 dataset. The upper part presents the results of single models and the lower part presents those that aggregate multiple models.
``Ours'' and ``Ours (pre)'' denote our framework without and with pretraining on the COCO dataset. ``Ours (pre \& fusion)'' denotes the results when fusing our two scale results.
The best and second-best results are highlighted in {\color{red}{red}} and {\color{blue}{blue}}, respectively. Best viewed in color.}
\label{table:comparision_voc12}
\end{table*}

\subsection{Results on Multi-Label Image Recognition}
In this subsection, we provide the comparison of the proposed framework with current state-of-the-art methods on the multi-label image recognition task.

\subsubsection{Comparison on Microsoft COCO}
We first present the evaluation results on the Microsoft COCO \cite{lin2014microsoft} dataset. To fairly compare the OP, OR, OF1 and CP, CR, CF1 metrics with the top-3 constraint, we follow the existing method \cite{wang2016cnn} to exclude labels with probabilities below a threshold (0.5 in our experiments). The comparison results are presented in Table \ref{table:result-coco}. As shown, the current best-performing methods are ResNet-SRN and ML-GCN, in which ResNet-SRN builds on ResNet-101 and applies an attention mechanism to model the label relations, while ML-GCN further uses a graph convolutional network to capture the label dependencies and achieves a mAP of 83.1\%. In contrast to these methods, our framework models label dependencies in both the feature and label spaces, leading to notable performance improvements on all the metrics. Specifically, it achieves a mAP of 84.3\%, improving on the results of the previous best methods by 1.2\%.

\begin{figure*}[!t]
\centering
\includegraphics[width=0.98\linewidth]{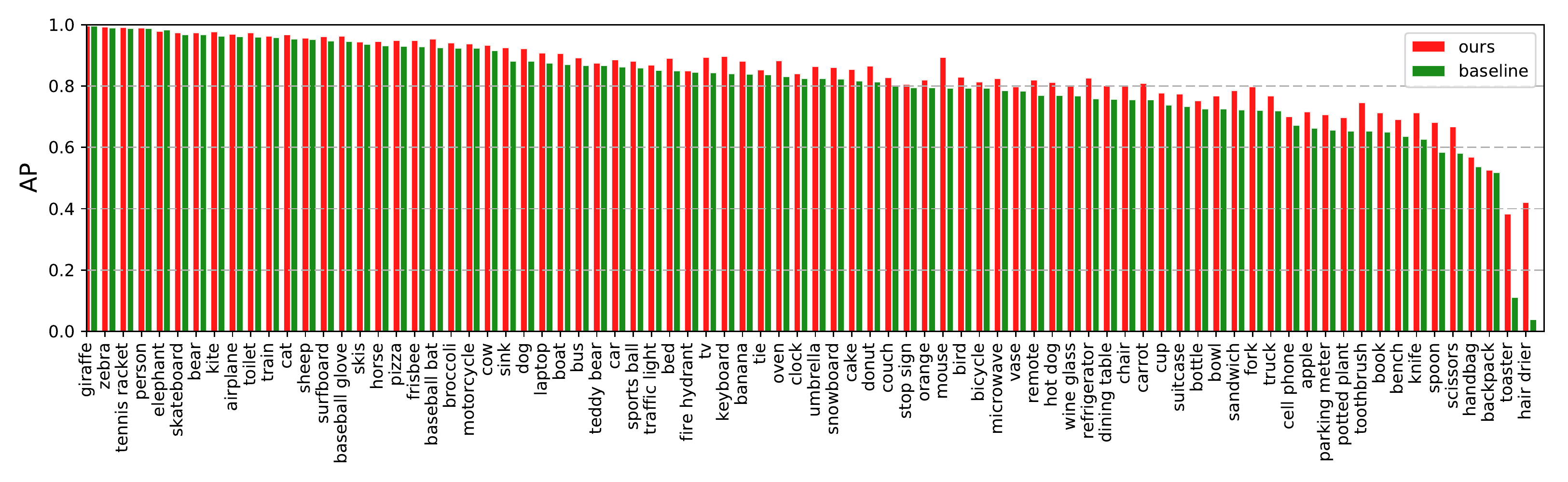} 
\caption{The AP (in \%) of each category of our proposed framework and the ResNet-101 baseline on the Microsoft COCO dataset.}
\label{fig:ap-comparison}
\end{figure*}

\begin{table*}[!t]
\centering
\begin{tabular}{c|c|cccccc|cccccc}
\hline
& & \multicolumn{6}{c|}{Top 3} &\multicolumn{6}{c}{All} \\
\hline
\centering Methods & mAP & CP & CR & CF1 & OP & OR & OF1 & CP & CR & CF1 & OP & OR & OF1 \\
\hline
\hline
ResNet-101 \cite{he2016deep} & 30.9 & 39.1 & 25.6 & 31.0 & 61.4 & 35.9 & 45.4 & 39.2 & 11.7 & 18.0 & 75.1 & 16.3 & 26.8 \\
ML-GCN \cite{DBLP:conf/cvpr/ChenWWG19} & 32.6 & 42.8 & 20.2 & 27.5 & 66.9 & 31.5 & 42.8 & 39.4 & 10.6 & 16.8 & 77.1 & 16.4 & 27.1 \\
\hline
Ours & \textbf{37.4} & \textbf{47.4} & \textbf{24.7} & \textbf{32.5} & \textbf{66.9} & \textbf{36.5} & \textbf{47.2} & \textbf{48.7} & \textbf{12.1} & \textbf{19.4} & \textbf{78.6} & \textbf{17.1} & \textbf{28.1} \\
\hline
\end{tabular}
\caption{Comparison of the mAP, CP, CR, CF1 and OP, OR, OF1 (in \%) scores of our framework and other state-of-the-art methods under settings of all and top-3 labels on the VG-500 dataset.}
\vspace{-10pt}
\label{table:result-vg}
\end{table*}

\subsubsection{Comparisons on Pascal VOC 2007 and 2012}
Here, we present the AP of each category and the mAP over all categories on the Pascal VOC 2007 dataset in Table \ref{table:comparision_voc07}. Most of the existing state-of-the-art methods focus on locating informative regions (e.g., proposal candidates \cite{yang2016exploit,wei2016hcp,zhang2018multi}, attentive regions \cite{wang2017multi}, or random regions \cite{wang2016beyond}) to aggregate local discriminative features to facilitate recognizing multiple labels in the given image. For example, RCP achieves a mAP of 92.5\%, which is the best result to date. In contrast, our framework incorporates category semantics to better learn semantic-specific features and explores their interactions under the explicit guidance of statistical label dependencies, which further improves the mAP to 93.6\%. In addition, when our framework is pretrained on the COCO dataset, it achieves an even better performance, i.e., 95.0\%, as shown in Table \ref{table:comparision_voc07}. Note that the existing methods aggregate multiple models \cite{simonyan2015very} or fuse the results with other methods \cite{yang2016exploit} to improve the overall performance. For example, FeV+LV (fusion) aggregates its results with those of VGG16\&19+SVM, which improves the mAP from 90.6\% to 92.0\%. Although our results are generated by a single model, it still outperforms all these aggregated results.

We also compare the performances on the Pascal VOC 2012 dataset, as depicted in Table \ref{table:comparision_voc12}. VOC 2012 is more challenging and larger in size, but our framework still achieves the best performance compared with the state-of-the-art competitors. Specifically, it obtains mAP scores of 93.9\% and 95.0\% without and with pretraining on the COCO dataset, respectively, with improvements over the previous best method by 1.7\% and 2.8\%, respectively. Similarly, the existing methods also aggregate the results of multiple models to boost performance. To ensure a fair comparison, we trained another model with an input of $448 \times 448$. Specifically, during training, we resized the input image to $512 \times 512$, and randomly choose a number from {512, 448, 384, 320, 256} as the width and height to randomly crop patches; then, we further resized the cropped patches to $448 \times 448$. We denote the original model as ``scale 640" and this modified model as ``scale 512". The two models are both pretrained on the COCO dataset and retrained on the VOC 2012 dataset. Then, we perform ten crop evaluations (the four corner crops and the center crop as well as their horizontally flipped versions) for each scale and aggregate the results from the two scales. As shown in the lower part of Table \ref{table:comparision_voc12}, our framework boosts the mAP to 95.6\%, outperforming all the existing methods with single and multiple models.

\subsubsection{Comparison on Visual Genome 500}
The Visual Genome 500 is a new dataset that can be used to evaluate multi-label image recognition with larger-scale categories; thus, no existing methods have reported their results on this dataset. To demonstrate the effectiveness of our proposed framework on this dataset, we implemented a ResNet-101 baseline network and trained it using the same process as was used for our model previously. Because ML-GCN \cite{chen2019multi} is the best-performing method on the Microsoft-COCO dataset, we further follow its released code to train the model on VG-500 for comparison. All the methods were trained on the training set and evaluated on the test set. The comparison results are presented in Table \ref{table:result-vg}. Our framework performs considerably better than the existing state-of-the-art and the ResNet-101 baseline methods on all metrics. Specifically, it achieves the mAP, top-3 CF1 and OF1, and top-all CF1 and OF1 of 37.4\%, 32.5\%, 47.2\%, 19.4\%, 28.1\%, improving the existing best method ML-GCN by 4.8\%, 1.5\%, 1.8\%, 1.4\%, 1.0\%, respectively. This comparison suggests that our framework also performs better for recognizing large-scale categories.

\subsection{Results on Multi-Label Few-Shot Learning}
In this subsection, we present comparisons of our proposed framework with current methods on the multi-label few-shot learning task.

\subsubsection{Comparison on Microsoft COCO}
The previous best-performing method for multi-label few-shot learning is LaSO \cite{alfassy2019laso}, and its results are mainly presented on Microsoft COCO. In this section, we compare our proposed method with this work and the baseline method that uses the mixUp \cite{zhang2017mixup} augmentation technique to directly train on the small novel set. The results are presented in Table \ref{table:result-fs-coco}. As shown, LaSO achieves mAPs of 45.3\% and 58.1\% on the 1-shot and 5-shot settings, respectively. In contrast, our method incorporates prior knowledge of category correlation to guide feature and semantic propagation among the different categories, leading to superior performance. Specifically, our method's mAP scores on the 1-shot and 5-shot settings are 52.3\% and 63.5\%, outperforming LaSO by 7.0\% and 5.4\%, respectively. Note that we use ResNet-101 \cite{he2016deep} as the backbone, while LaSO uses GoogleNet-v3 \cite{szegedy2015going,szegedy2016rethinking}. To ensure a fair comparison, we further replaced our backbone network with GoogleNet-v3, keeping the other components and the training process unchanged. Nevertheless, our framework still outperforms LaSO, achieving mAP scores of 49.4\% and 61.0\% on the 1-shot and 5-shot settings, respectively.

\begin{table}[!t]
\centering
\begin{tabular}{c|cc}
\hline
\centering Methods & 1-shot & 5-shot \\
\hline 
\hline
Baseline w/ Mixup Aug \cite{zhang2017mixup} & 40.2 & 54.0 \\
LaSO (GoogleNet-v3) \cite{alfassy2019laso} & 45.3 & 58.1 \\
\hline
Ours (GoogleNet-v3) & 49.4 & 61.0 \\
Ours (ResNet-101) & 52.3 & 63.5\\
\hline
\end{tabular}
\vspace{2pt}
\caption{Comparison of the mAP (in \%) on 1-shot and 5-shot settings on the Microsoft COCO dataset. We present the results of both 1-shot and 5-shot settings.}
\label{table:result-fs-coco}
\end{table}

\begin{figure*}[!t]
\centering
\includegraphics[width=0.9\linewidth]{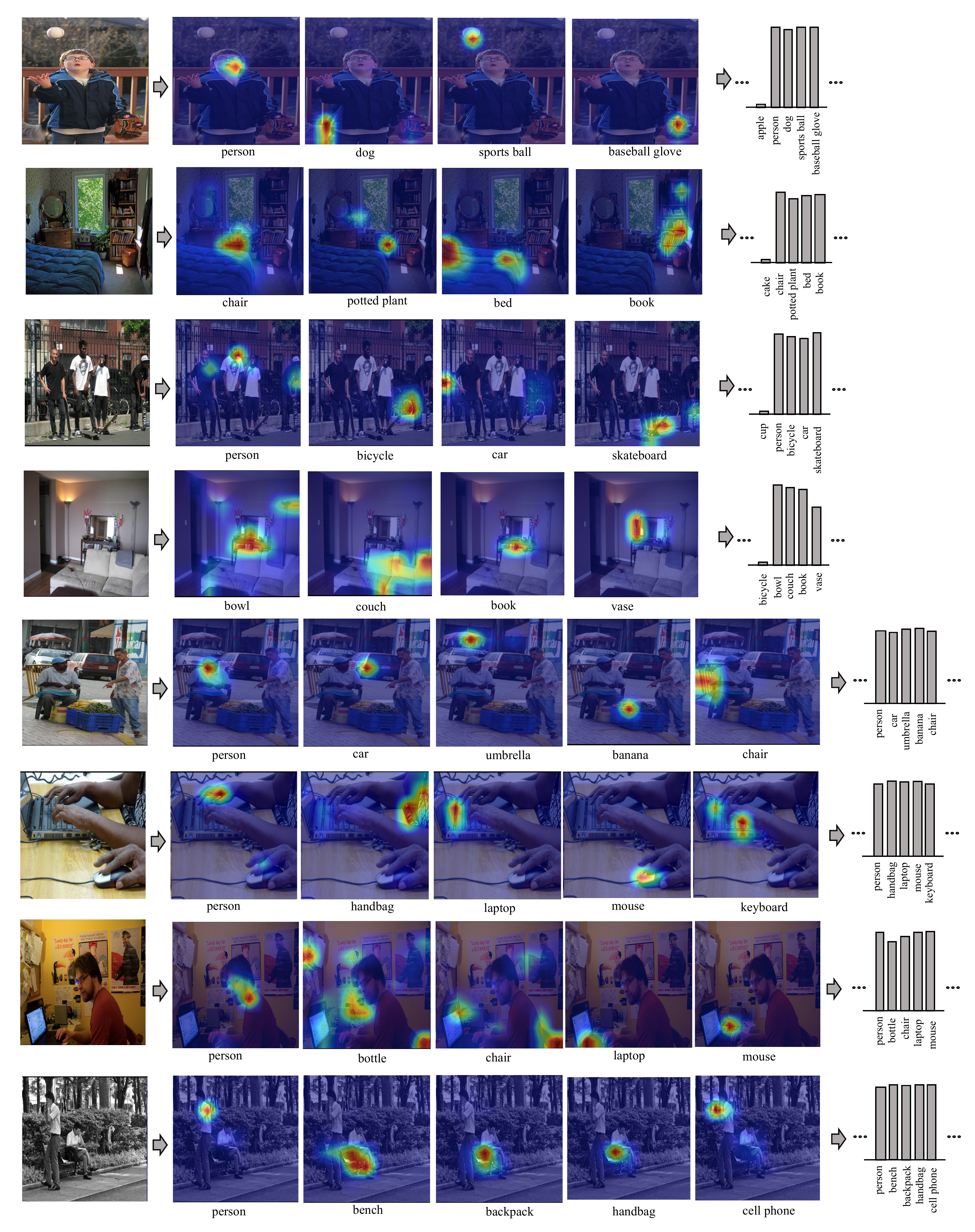} 
\caption{Several examples of input images (left), semantic feature maps corresponding to categories with the top-$k$ ($k=4,5$) highest confidences (middle), and the predicted label distribution (right).}
\label{fig:semantic-map}
\vspace{4pt}
\end{figure*}

\subsubsection{Comparison on Visual Genome 500}
Microsoft-COCO contains only 80 categories, but current works \cite{zhang2019few,chen2020knowledge} for few-shot learning may contain thousands of categories, which is more realistic. To remedy this issue, we utilize Visual Genome 500 as a new evaluation benchmark. Because LaSO is the previous best-performing method, we followed the code \footnote{https://github.com/leokarlin/LaSO} released by the authors to train the LaSO model on the VG-500 dataset. For a fair comparison, we used ResNet-101 as the backbone network for both methods. The results are listed in Table \ref{table:result-fs-vg}. Our framework clearly outperform the LaSO by a sizable margin. Specifically, our framework achieves mAPs of 20.7\% and 26.1\% on the 1-shot and 5-shot settings, outperforming LaSO by 4.1\% and 4.3\%, respectively.

\begin{table}[ht]
\centering
\begin{tabular}{c|cc}
\hline
\centering Methods & 1-shot & 5-shot \\
\hline 
\hline
LaSO \cite{alfassy2019laso} & 16.6 & 21.8 \\
\hline
Ours & 20.7 & 26.1\\
\hline
\end{tabular}
\vspace{2pt}
\caption{Comparison of the mAP (in \%) on 1-shot and 5-shot settings on the VG-500 dataset. We present the results of 1-shot and 5-shot settings.}
\label{table:result-fs-vg}
\end{table}

\subsection{Ablative study}
The proposed framework builds on ResNet-101 \cite{he2016deep}; thus, we first compare it with this baseline to analyze the contributions of knowledge-guided graph routing (KGGR). Specifically, we simply replace the last fully connected layer of ResNet-101 with a 2,048-to-$C$ fully connected layer and use $C$ sigmoid functions to predict the probability of each category. The training and test settings are the same as those described in Section \ref{sec:mlr}. We conducted the experiments on the Microsoft-COCO dataset and present the results in Table \ref{table:result-ablation-mlr}. As shown, the mAP drops from 84.3\% to 80.3\%. To more deeply analyze the performance comparisons, we further present the AP of each category in Figure \ref{fig:ap-comparison}, which shows that the AP improvement by our framework is more evident for the categories that are more difficult to recognize (i.e., the categories for which the baseline obtains lower AP). For example, for categories such as giraffe and zebra, the baseline obtains a very high AP; thus, our framework achieves only slight improvements. In contrast, for more difficult categories such as toaster and hair drier, our framework improves the AP by a sizable margin---27.3\% and 38.3\% improvements for toaster and hair drier, respectively.

We also present a comparison with the baseline ResNet-101 on the multi-label few-shot learning task in Table \ref{table:result-ablation-mlfsl}. As shown, our method achieves evident improvements over the baseline (i.e., improving the mAP by 8.8\% and 7.0\% on the 1-shot and 5-shot settings, respectively).

\begin{table}[h]
\centering
\begin{tabular}{c|c}
\hline
\centering Methods & mAP \\
\hline
\hline
ResNet-101 \cite{he2016deep} &80.3\\
\hline
Ours w/o KEFP & 80.9\\
Ours w/o SGA & 82.2\\
Ours w/o GFP & 80.6\\
Ours w/o GSP & 83.8 \\
Ours & 84.3 \\
\hline
\end{tabular}
\vspace{2pt}
\caption{Comparison of mAP (in \%) of our framework (Ours), our framework without the graph feature propagation module (Ours w/o GFP), our framework without the graph semantic propagation module (Ours w/o GSP), our framework without semantically guided attention (Ours w/o SGA), our framework without knowledge-embedded feature propagation (Ours w/o KEFP), and the baseline ResNet-101 from the multi-label image recognition task on the Microsoft-COCO dataset.}
\label{table:result-ablation-mlr}
\end{table}

\begin{table}[h]
\centering
\begin{tabular}{c|c|c}
\hline
\centering Methods & 1-shot & 5-shot \\
\hline
\hline
ResNet-101 \cite{he2016deep} & 43.5 & 56.5\\
\hline
Ours w/o GFP & 44.1 & 56.0  \\
Ours w/o GSP & 50.4 & 61.0 \\
Ours & 52.3 & 63.5 \\
\hline
\end{tabular}
\vspace{2pt}
\caption{Comparison of mAP (in \%) of our framework (Ours), our framework without the graph feature propagation module (Ours w/o GFP), our framework without the graph semantic propagation module (Ours w/o GSP), and the baseline ResNet-101 from the multi-label few-shot task on the Microsoft-COCO dataset.}
\label{table:result-ablation-mlfsl}
\end{table}

The foregoing comparisons verify the contribution of the proposed KGGR as a whole. Actually, the KGGR contains two graph propagation modules that conduct information interactions to learn features and classifiers. In the following, we further conduct ablation experiments to analyze the true contribution of each module.

\begin{figure}[!t]
\centering
\includegraphics[width=0.90\linewidth]{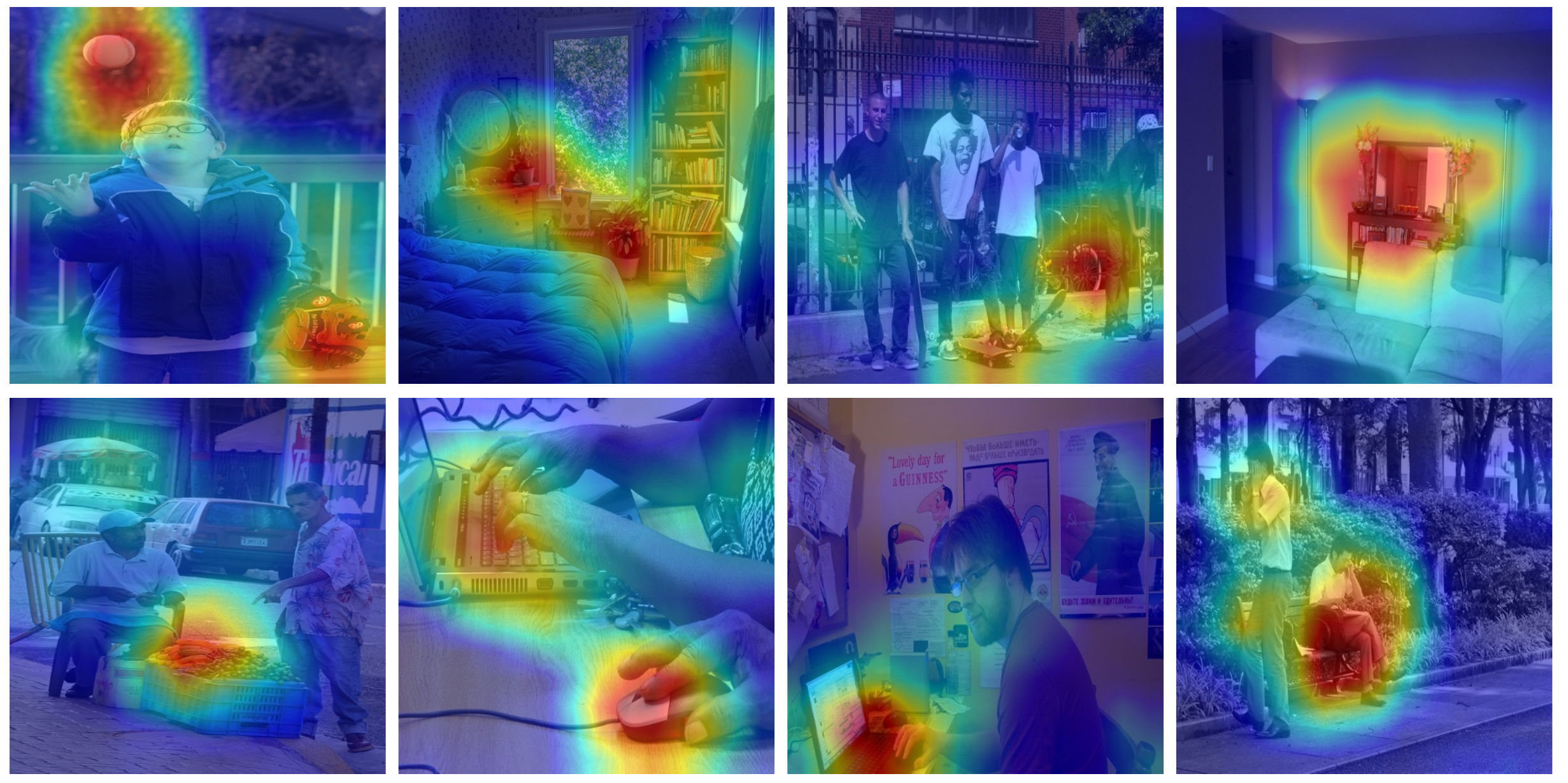} 
\caption{Several examples of the feature maps generated by the baseline ResNet-101. The samples are the same as those in Figure \ref{fig:semantic-map} for direct comparisons.}
\label{fig:visualization-baseline}
\end{figure}

\subsubsection{Contribution of graph feature propagation}
Graph feature propagation can help learn the contextualized feature vectors for each category, and it is the key module in this work. Here, we first remove this module and directly use $\mathbf{f}$ for classification. We find that the resulting model suffers from an obvious performance drop (i.e., the mAP decreased from 84.3\% to 80.6\%). Moreover, this module uses a semantically guided attention mechanism that decouples the image into C category-specific feature vectors and a knowledge-embedded feature propagation that explores feature interactions to learn contextualized features. We also analyze these two components in the following. 1) To analyze the semantically guided attention mechanism, we remove it and directly use $\mathbf{f}$ to initialize the graph nodes. As shown in Table \ref{table:result-ablation-mlfsl}, this model achieve a mAP of 80.9\%. It performs slightly better than the baseline method, because it does not incur any additional information but does increase the model complexity. As discussed above, this component can learn semantic-specific feature maps that focus on corresponding semantic regions via semantically guided attention mechanism. We present some examples in Figure \ref{fig:semantic-map}. From left to right in each row are the input images, the semantic maps corresponding to categories with the top $k (k=4,5)$ highest confidences, and the predicted label distribution. These results show that our semantically guided attention mechanism is able to highlight the semantic regions well when the objects of the corresponding categories exist. Taking the first sample as an example, it contains person, dog, sport ball, and baseball glove objects, which highlight the corresponding regions of these semantic objects, respectively. Similar phenomena can be observed for the other examples. To clearly verify that it is the semantically guided attention mechanism that brings about this appealing characteristic, we further visualize the feature maps generated by the baseline ResNet-101. As shown in Figure \ref{fig:visualization-baseline}, it tends to highlight the most salient regions, but neglect some regions that are less salient but equally important, e.g., the dog in the first example and the chair in the seventh example.
2) To validate the knowledge-embedded feature propagation component, we remove the graph propagation network; thus, $\mathbf{o}_c=\mathbf{f}_c$. As shown in Table \ref{table:result-ablation-mlr}, the mAP is 82.2\%, a decrease in the mAP of 2.1\%.

We also analyze the effect of this module on the multi-label few-shot task and present the result with and without the graph feature propagation component in Table \ref{table:result-ablation-mlfsl}. The mAP falls from 52.3\% and 63.5\% to 44.1\% and 56.0\% on the 1-shot and 5-shot settings, respectively.

\subsubsection{Contribution of graph semantic propagation}
Graph semantic propagation helps transfer the information of correlated categories to better learn the classifiers. To validate its contribution, we remove this component and use $C$ fully connected layers that directly take the corresponding contextualized feature vector as input to predict the probability of the corresponding category. As shown in Table \ref{table:result-ablation-mlr}, the mAP decreases from 84.3\% to 83.8\% on Microsoft COCO. We found a similar phenomenon for the multi-label few-shot task, where the mAP decreases from 52.3\% and 63.5\% to 50.4\% and 61.0\% on the 1-shot and 5-shot settings, respectively.

\subsubsection{Analysis of different losses}
In this work, we learn a contextualized feature vector for each category and adopt the logistic regressor to predict the existence probability for each category. We follow recent multi-label works \cite{chen2019multi,Tokmakov2019ICCV} to train the models with the cross-entropy loss. Earlier works \cite{wei2016hcp,chen2018recurrent} has also used the euclidean loss between the output score and normalized ground truth to train the model. In this part, we conduct an experiment that follows works \cite{wei2016hcp,chen2018recurrent} to replace the loss with Euclidean loss to analyze the effect of different losses. As shown in Table \ref{table:result-ablation-loss-mlr}, the model trained with the entropy loss achieves much better performance.

\begin{table}[h]
\centering
\begin{tabular}{c|c}
\hline
\centering Methods & mAP \\
\hline
\hline
Ours EU & 79.2 \\
Ours CE& 84.3 \\
\hline
\end{tabular}
\vspace{2pt}
\caption{Comparison of mAP (in \%) of our framework with the cross entropy (Ours CE) and euclidean (Ours EU) losses on the Microsoft-COCO dataset.}
\label{table:result-ablation-loss-mlr}
\end{table}


\section{Conclusion}
In this work, we explore integrating prior knowledge of label correlations into deep neural networks to guide learning both feature and classifier representations. To achieve this end, we propose a novel knowledge-guided graph routing (KGGR) framework that consists of two graph propagation mechanisms. The first propagation mechanism introduces category semantic to guide learning semantic-specific features and exploit a graph neural network to explore feature interaction to learn contextualized feature representation for each category. The second propagation mechanism exploits another graph neural network that takes initial classifier weight as input and transfers classifier information through different categories to help better learn the classifiers. We apply the proposed framework to both multi-label image recognition and multi-label few-shot learning tasks on the Microsoft-COCO, Pascal VOC 2007 \& 2012, and Visual Genome datasets, and demonstrate its effectiveness over all existing leading methods on both two tasks.

\ifCLASSOPTIONcaptionsoff
\newpage
\fi



%

{
\bibliographystyle{IEEEtran}
\bibliography{reference}
}



%

\begin{IEEEbiography}[{\includegraphics[width=1in,height=1.25in,clip,keepaspectratio]{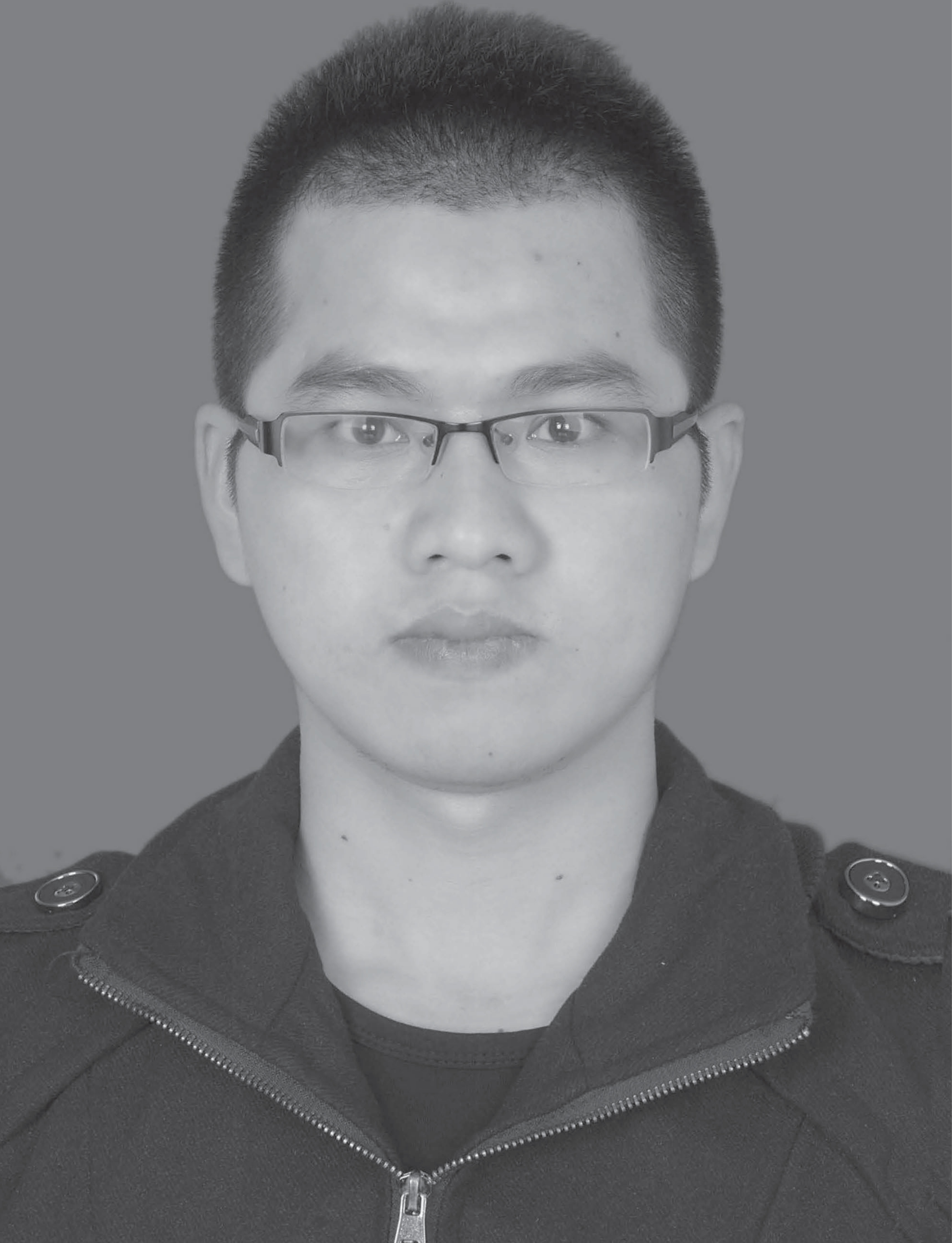}}]{Tianshui Chen} received a Ph.D. degree in computer science at the School of Data and Computer Science Sun Yat-sen University, Guangzhou, China, in 2018. Before that, he received a B.E. degree from the School of Information and Science Technology. He is currently an Associate Research Director at DMAI Co., Ltd. His current research interests include computer vision and machine learning. He has authored and coauthored more than 20 papers published in top-tier academic journals and conferences. He has served as a reviewer for numerous academic journals and conferences, including TPAMI, TIP, TMM, TNNLS, CVPR, ICCV, ECCV, AAAI, and IJCAI. He was the recipient of the Best Paper Diamond Award at IEEE ICME 2017.
\end{IEEEbiography}

\begin{IEEEbiography}[{\includegraphics[width=1in,clip]{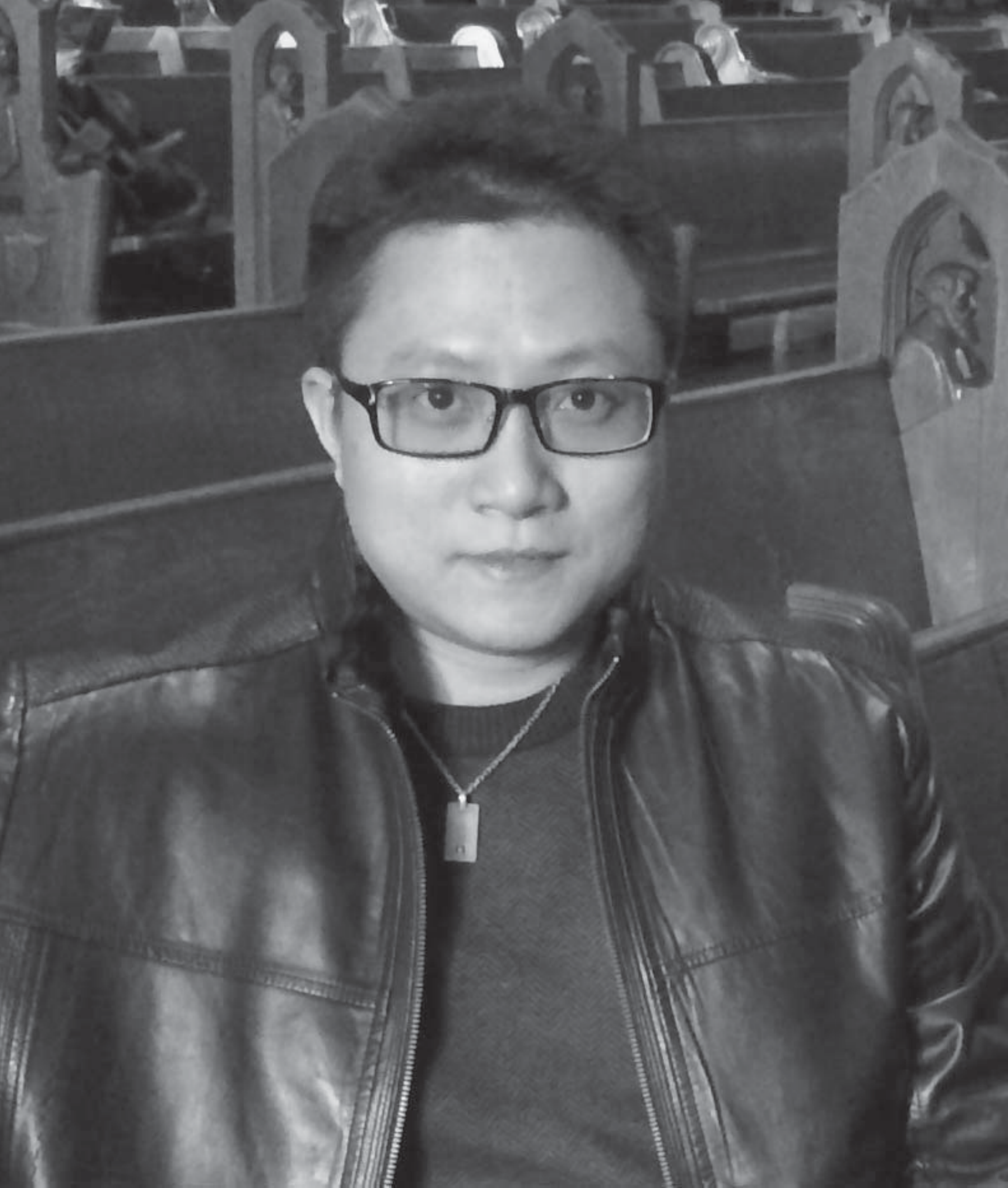}}]{Liang Lin} (M'09, SM'15) is a Full Professor of computer science at Sun Yat-sen University. He has authored or co-authored more than 200 papers in top-tier academic journals and conferences with more than 12,000 citations. He is an associate editor of IEEE Trans. Human-Machine Systems and IET Computer Vision. He served as Area Chairs for numerous conferences such as CVPR, ICCV, and IJCAI. He is the recipient of numerous awards and honors including Wu Wen-Jun Artificial Intelligence Award, CISG Science and Technology Award, ICCV Best Paper Nomination in 2019, Annual Best Paper Award by Pattern Recognition (Elsevier) in 2018, Best Paper Dimond Award in IEEE ICME 2017, Google Faculty Award in 2012, and Hong Kong Scholars Award in 2014. He is a Fellow of IET.
\end{IEEEbiography}

\begin{IEEEbiography}[{\includegraphics[width=1in,height=1.25in,clip,keepaspectratio]{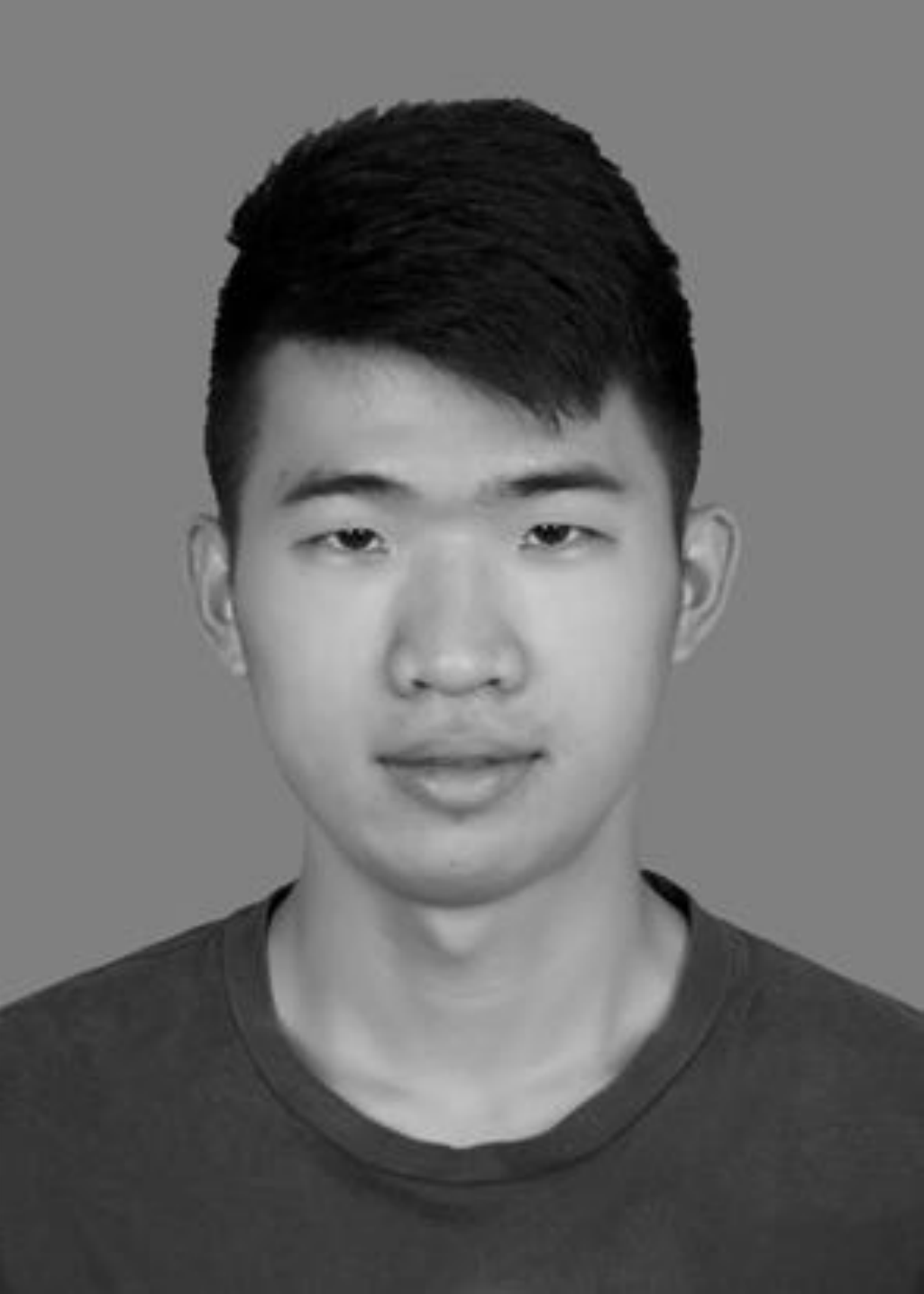}}]{Riquan Chen}
received a B.E. degree from the School of Mathematics, Sun Yat-sen University, Guangzhou, China, in 2017, where he is currently pursuing his master's degree in computer science at the School of Data and Computer Science. His current research interests include computer vision and machine learning.
\end{IEEEbiography}

\begin{IEEEbiography}[{\includegraphics[width=1in,height=1.25in,clip,keepaspectratio]{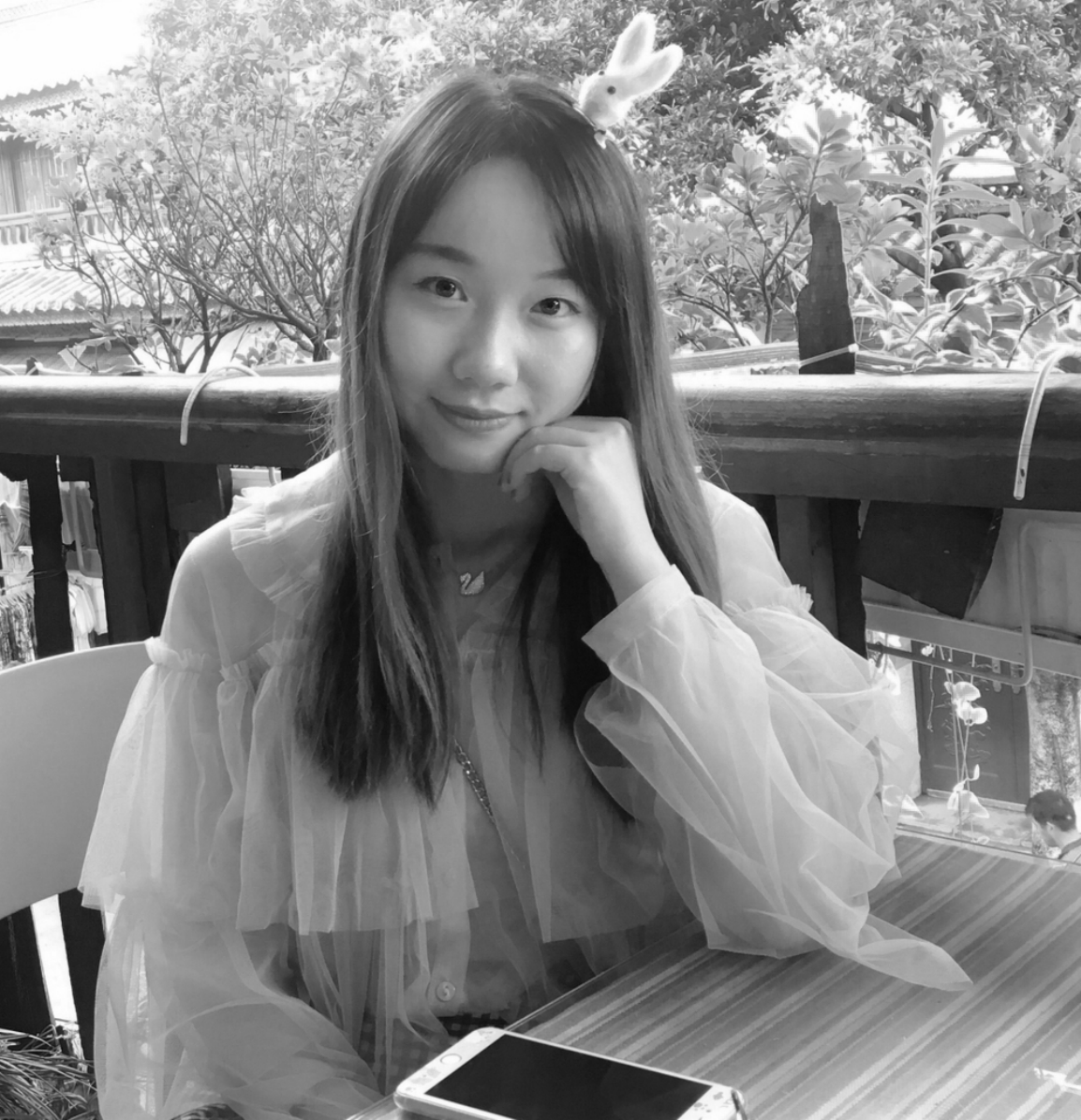}}]{Xiaolu Hui} received her B.E. degree from the School of Computing, Northwestern Polytechnical University, Shaanxi, China, in 2018. She is currently pursuing a master’s degree in computer science with the School of Data and Computer Science at Sun Yat-sen University. Her current research interests include computer vision and machine learning.\end{IEEEbiography}

\begin{IEEEbiography}[{\includegraphics[width=1in,height=1.25in,clip,keepaspectratio]{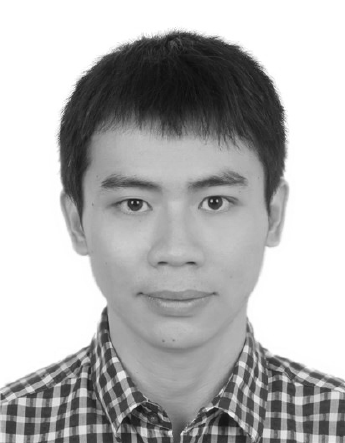}}]{Hefeng Wu}
received a B.S. degree in computer science and technology and a Ph.D. in computer application technology from Sun Yat-sen University, China, in 2008 and 2013, respectively. He is currently a full research scientist with the School of Data and Computer Science, Sun Yat-sen University, China. His research interests include computer vision, multimedia, and machine learning.
\end{IEEEbiography}

\end{document}